%% file: main.tex
\documentclass{article}

\usepackage[preprint]{neurips_data_2024}
\usepackage{latexsym}
\usepackage{microtype}
\usepackage{amssymb}
\usepackage{mathtools}
\usepackage{nicematrix}
\usepackage{colortbl}
\usepackage{hyperref}
\usepackage{url}
\usepackage{amsfonts}
\usepackage{verbatim}
\usepackage{caption}
\usepackage{longtable}
\usepackage{supertabular}
\usepackage{float}
\usepackage{tabularx}
\usepackage{amsmath}
\usepackage{graphicx}
\usepackage{xcolor}
\usepackage{multirow}
\usepackage{makecell}
\usepackage{arydshln}
\usepackage{booktabs}
\usepackage{xspace}
\usepackage{wrapfig}
\usepackage[commandnameprefix=ifneeded]{changes}
\usepackage{enumitem}
\usepackage{xspace}
\usepackage{textcomp}
\usepackage{makecell}
\usepackage{multicol}
\usepackage{lscape} 
\usepackage{siunitx}
\usepackage{algorithm}
\usepackage{algpseudocode}
\usepackage{stfloats}
\usepackage{amssymb}
\usepackage{pifont}
\definecolor{deepgreen}{HTML}{006400}
\usepackage{scrextend}
\usepackage{threeparttable}
\usepackage{colortbl}
\usepackage{listings}
\usepackage{subfiles}
\usepackage{array}
\usepackage{latexsym}
\usepackage[T1]{fontenc}
\usepackage[utf8]{inputenc}
\usepackage{microtype}
\usepackage{xurl} 
\usepackage{nicefrac}       
\usepackage{changepage}
\usepackage{xargs}
\usepackage{wrapfig,lipsum,booktabs}
\usepackage{longtable}
\usepackage{subcaption}
\usepackage{pgfplots}
\usetikzlibrary{pgfplots.groupplots}
\pgfplotsset{compat=1.3}
\usepackage{tikz}
\usetikzlibrary{patterns}
\usepackage{placeins} 

\usepackage[T1]{fontenc}

\usepackage[utf8]{inputenc}

\usepackage{microtype}

\usepackage{inconsolata}
\newcommand{\yes}{{\color{deepgreen}\ding{51}}}
\definecolor{ao(english)}{rgb}{0.0, 0.5, 0.0}
\newcommand{\no}{{\color{red}\ding{55}}}

\title{
\raisebox{-3pt}{\includegraphics[height=1.6em]{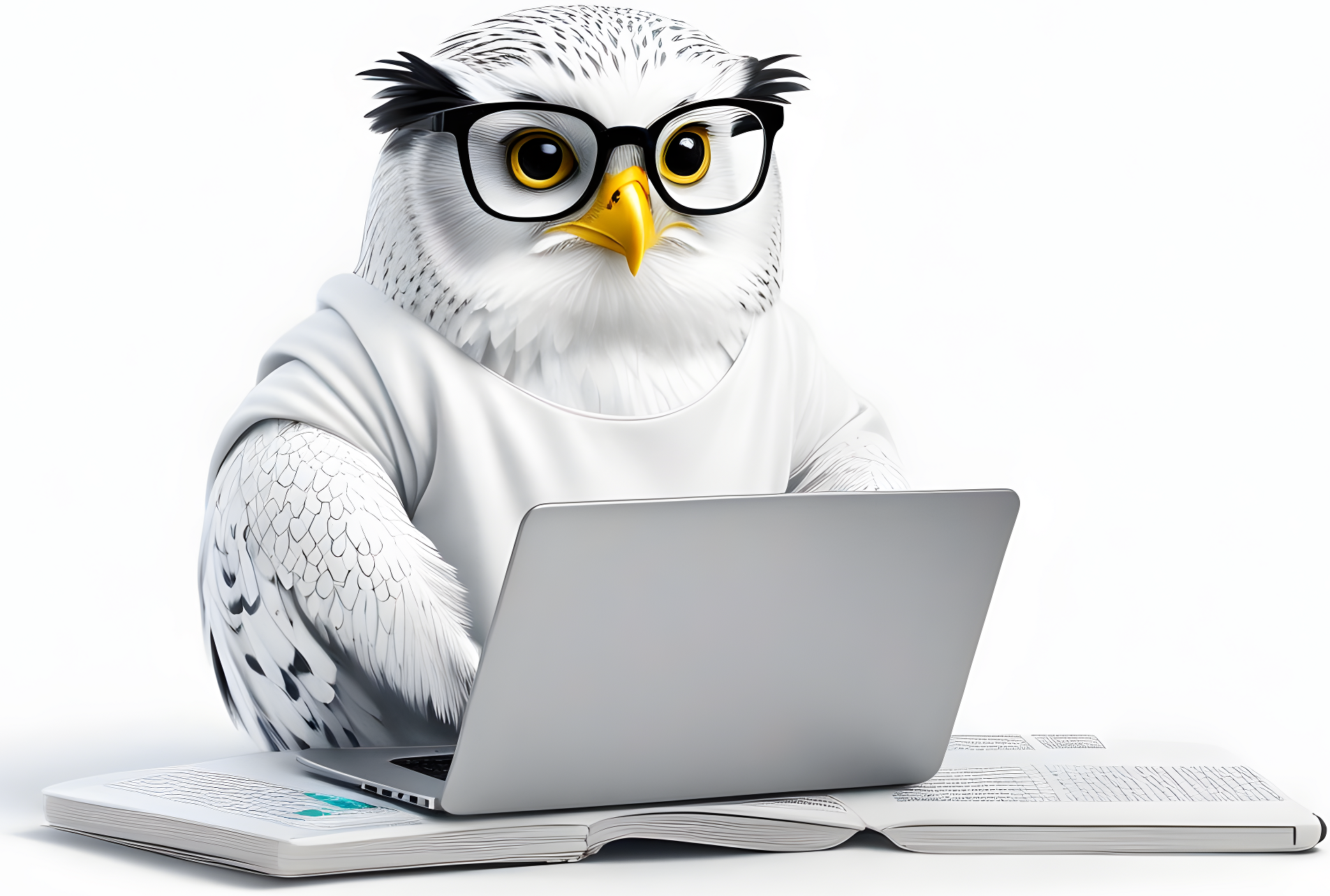}}\xspace
\benchname: Evaluating Large Language Models and Agents for Machine Learning Tasks on Repository-Level Code}

\makeatletter
\AtBeginDocument{%
  \renewcommand{\sectionautorefname}{\S\@gobble}

  \renewcommand{\subsectionautorefname}{\S\@gobble}
}
\makeatother

\usepackage{listings}
\usepackage{xcolor}
\newcommand{\benchname}{\textsc{ML-Bench}\xspace}

\newcommand{\huggingface}{\raisebox{-2.pt}{\includegraphics[height=1.05em]{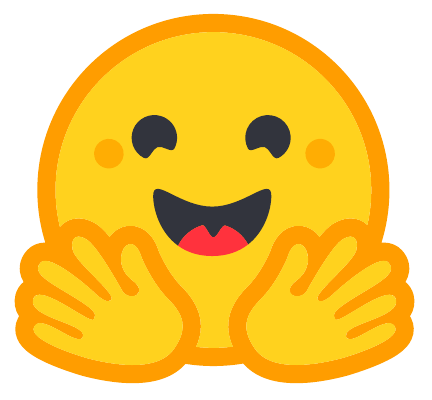}}\xspace}
\newcommand{\slack}{\raisebox{-3.5pt}{\includegraphics[height=1.05em]{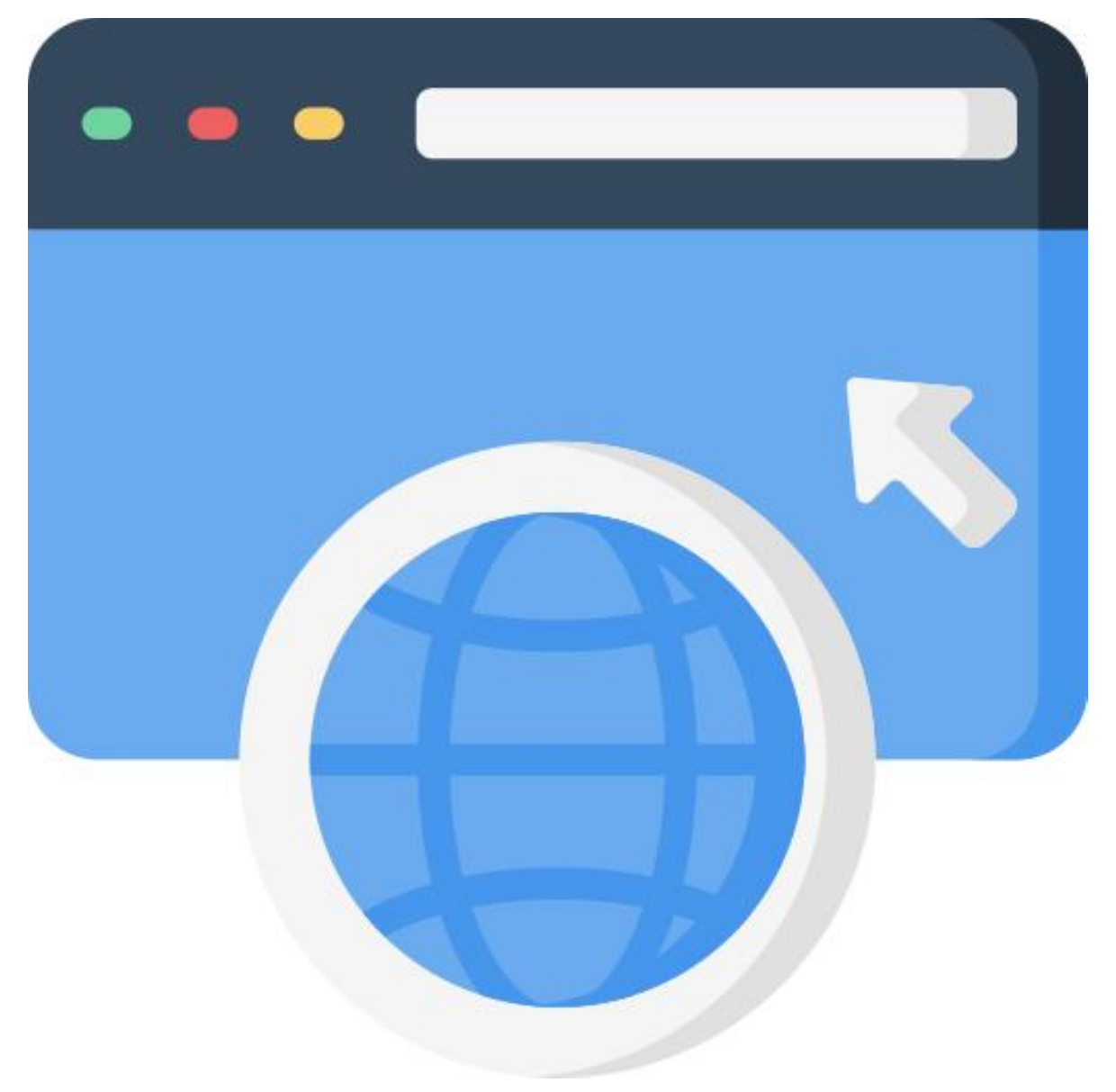}}\xspace}

\newcommand{\github}{\raisebox{-1.5pt}{\includegraphics[height=1.05em]{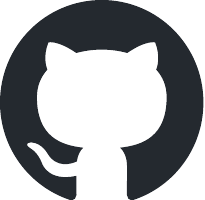}}\xspace}

\author{
\vspace{2.5mm} 
\hspace{-6.5mm}
        Xiangru Tang$^{*\spadesuit}$ \hspace{2.5mm}
        Yuliang Liu\thanks{Contribute equally.}$^{*\clubsuit}$\hspace{2.5mm}
        Zefan Cai$^{*\heartsuit}$\hspace{2.5mm}
        Yanjun Shao$^{*\spadesuit}$\hspace{3mm}
        Junjie Lu$^{\clubsuit}$\hspace{3mm}
        Yichi Zhang$^{\heartsuit}$\hspace{3mm}
          \vspace{-1mm}
    \\
    \textbf{\hspace{-5.5mm}
        Zexuan Deng$^{\clubsuit}$\hspace{3mm}
        Helan Hu$^{\heartsuit}$\hspace{3mm}
        Kaikai An$^{\heartsuit}$\hspace{3mm}
        Ruijun Huang\hspace{4mm}
        Shuzheng Si$^{\heartsuit}$\hspace{3mm}
         Sheng Chen$^{\clubsuit}$
    } \vspace{1.5mm}
    \\
 \textbf{\hspace{-5.5mm}    
        Haozhe Zhao$^{\heartsuit}$\hspace{3mm}
        Liang Chen$^{\heartsuit}$\hspace{3mm}
        Yan Wang\hspace{3mm}
                Tianyu Liu$^{\heartsuit}$\hspace{3mm}
                Zhiwei Jiang$^{\clubsuit}$\hspace{3mm}
                Baobao Chang$^{\heartsuit}$
    } \vspace{1.5mm}
  \\
    \textbf{\hspace{-5.5mm}
            Yin Fang\hspace{3mm}
        Yujia Qin\hspace{3mm}
        Wangchunshu Zhou\hspace{3mm}
     Yilun Zhao$^\spadesuit$ \hspace{3mm}
             Arman Cohan$^\spadesuit$ \hspace{3mm}
        Mark Gerstein$^\spadesuit$ 
    }
    \\
    \\
    \hspace{-5.5mm}  $^\spadesuit$  Yale University  \hspace{4mm}  $^\clubsuit$   Nanjing University \hspace{4mm}  $^\heartsuit$ Peking University
\\
\\
    {\tt \href{mailto:xiangru.tang@yale.edu}{\hspace{-4.5mm}xiangru.tang@yale.edu, pi@gersteinlab.org}}
}

\begin{document}
\maketitle
\thispagestyle{plain}


\vspace{-7mm}

\begin{abstract}

Despite Large Language Models (LLMs) like GPT-4 achieving impressive results in function-level code generation, they struggle with repository-scale code understanding (e.g., coming up with the right arguments for calling routines), requiring a deeper comprehension of complex file interactions. 
Also, recently, people have developed LLM agents that attempt to interact with repository code (e.g., compiling and evaluating its execution), prompting the need to evaluate their performance.  
These gaps have motivated our development of \textsc{ML-Bench}, a benchmark rooted in real-world programming applications that leverage existing code repositories to perform tasks. 
Addressing the need for LLMs to interpret long code contexts and translate instructions into precise, executable scripts, \textsc{ML-Bench} encompasses annotated \textbf{9,641} examples across \textbf{18} GitHub repositories, challenging LLMs to accommodate user-specified arguments and documentation intricacies effectively.
To evaluate both LLMs and AI agents, two setups are employed: 
\textsc{ML-LLM-Bench} for assessing LLMs' text-to-code conversion within a predefined deployment environment, and \textsc{ML-Agent-Bench} for testing autonomous agents in an end-to-end task execution within a Linux sandbox environment. 
Our findings indicate that while GPT-4o leads with a Pass@5 rate surpassing \textbf{50\%}, there remains significant scope for improvement, highlighted by issues such as hallucinated outputs and difficulties with bash script generation. 
Notably, in the more demanding \textsc{ML-Agent-Bench}, GPT-4o achieves a \textbf{76.47\%} success rate, reflecting the efficacy of iterative action and feedback in complex task resolution. 
Our resources, including code, data, and models, are publicly available under the MIT license.

\end{abstract}

\raggedbottom

\begin{center}
\begin{tabular}{rcl}
    \github & \textbf{\small{Code}} & \url{https://github.com/gersteinlab/ML-bench}\\
    \huggingface & \textbf{\small{Benchmark}} & \url{https://hf.co/datasets/super-dainiu/ml-bench}\\
    \slack & \textbf{\small{Webpage}} & \url{https://ml-bench.github.io/}\\
\end{tabular}
\end{center}

\input{files/1-intro}
\input{files/2-dataset}

\input{files/4-setup}

\input{files/3-agent_intro}
\input{files/5-experiments}

\input{files/6-conclusion}

\clearpage

\input{files/8-ethics+repro}



\bibliography{main}
\bibliographystyle{plainnat}
\clearpage

\input{files/9-appendix}



\end{document}

%% file: files/1-intro.tex
\section{Introduction}
\label{sec:intro} 



Large Language Models (LLMs) like GPT-4 have demonstrated remarkable prowess, reaffirming their state-of-the-art status on function-level code generation within widely recognized benchmarks such as HumanEval, MBPP, and APPS \citep{Austin2021ProgramSW, chen2021codex, Hendrycks2021MeasuringCC, Li2022}. Nevertheless, their capability wanes at the repository scale, which poses the multifaceted challenge of processing extensive contexts and navigating the intricate web of file interdependencies. 
This reveals the first gap in benchmarking - \textit{the absence of thorough assessments to test LLMs' abilities in complex repository-level deployment tasks}, such as interacting with functions in different files.
Moreover, the emergence of recent AI agents capable of executing and iteratively refining code marks a significant advancement~\citep{Shinn2023ReflexionAA,yang2024swe,Zhou2023AgentsAO}. 
Yet, a second gap persists in benchmarking such agents: \textit{the need for a rigorous evaluation environment} that can effectively measure an agent's skill in the initialization and utilization of software repositories.

\begin{figure}[!h]
\centering
\vspace{-.16cm}
\includegraphics[width=0.95\textwidth]{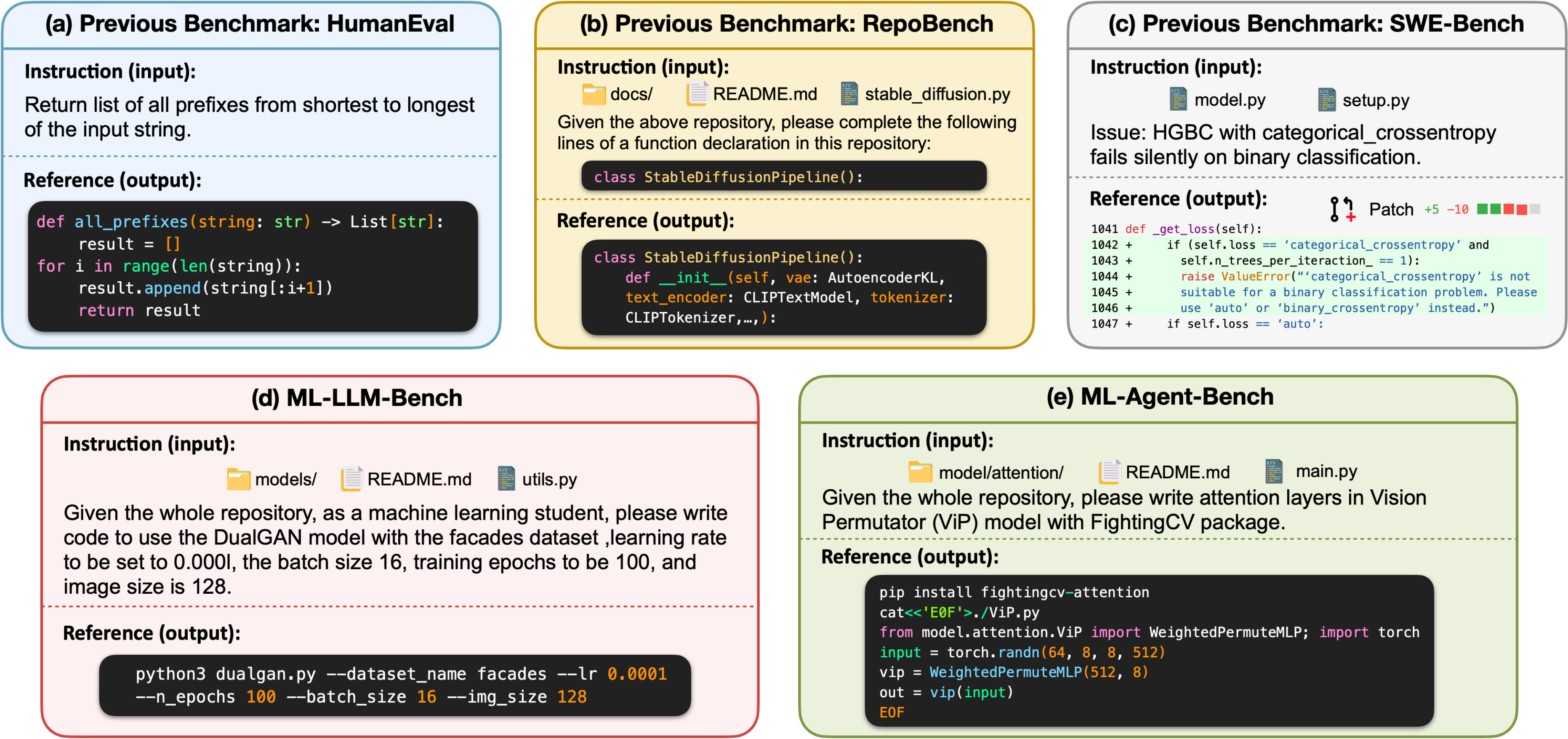}
\caption{\small Examples of \textsc{ML-Bench} compared with existing code benchmarks 
HumanEval \citep{chen2021codex}, 
RepoBench \citep{Liu2023RepoBenchBR}, and SWE-bench \citep{jimenezswe}. 
In \textsc{ML-Bench}, (1) models must \textbf{take repository-level code as input}, and (2) based on their understanding of the repository, models are required to compose \textbf{new code segments that do not exist} within the original repository.}
\label{fig:compare}
\vspace{-.2cm}
\end{figure}

\begin{wrapfigure}{R}{0.66\textwidth}
\centering 
\vspace{-0.5cm}
\includegraphics[width=0.62\textwidth]{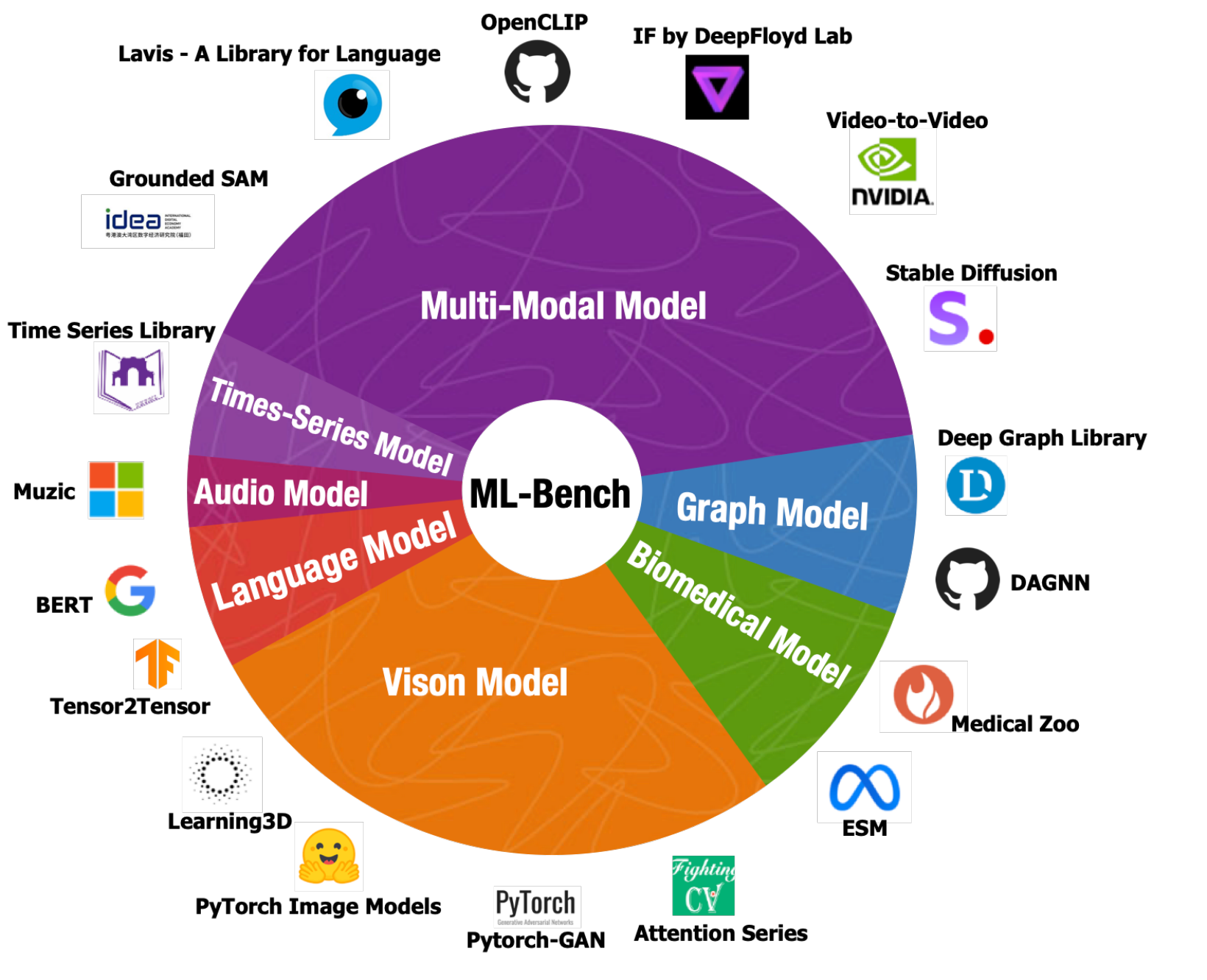}
\vspace{-0.2cm}
\caption{
\textsc{\footnotesize \benchname encompasses \textbf{18 prominent GitHub repositories} and it show the distribution of \textbf{9,641 samples}.} }
\label{fig:data_analysis}
\vspace{-0.25cm}
\end{wrapfigure}

We introduce \textsc{ML-Bench} based on common real-world programming workflows, often using existing repositories as libraries to accomplish tasks. 
\textsc{ML-Bench} seeks to test whether LLMs can generate executable code to invoke specific files or functions in a repository with appropriate arguments based on given instructions.
For instance, it might assess if an LLM can generate a command line to utilize \textit{$txt2img.py$} from an image generation model repository with parameters such as \textit{$ckpt$} to produce an image based on a text description, e.g.
python txt2img.py --prompt "a girl riding a horse" --ckpt SD2\_1\_v\_model.ckpt.
To effectively address these tasks, LLMs need to understand code at the repository level, which often involves processing a long context. In addition, they need to accurately configure parameters when executing program files.
For instance, LLMs must discern how to input values from user instructions correlate with flags found across code in various files, see Figure ~\ref{fig:compare}. Another crucial aspect of this process is understanding documentation—especially README files—which typically includes comprehensive instructions on how to employ the library, complete with task examples and arguments selection guidelines. In practice, user-specified arguments don't always match the examples provided in README files, thus requiring LLMs to modify arguments intelligently to meet specific user needs.
\textsc{ML-Bench} features \textbf{9,641} samples from \textbf{18} ML GitHub repositories, depicted in Figure \ref{fig:data_analysis}.

\begin{figure}[!h]
\centering
\vspace{-.16cm}
\includegraphics[width=0.95\textwidth]{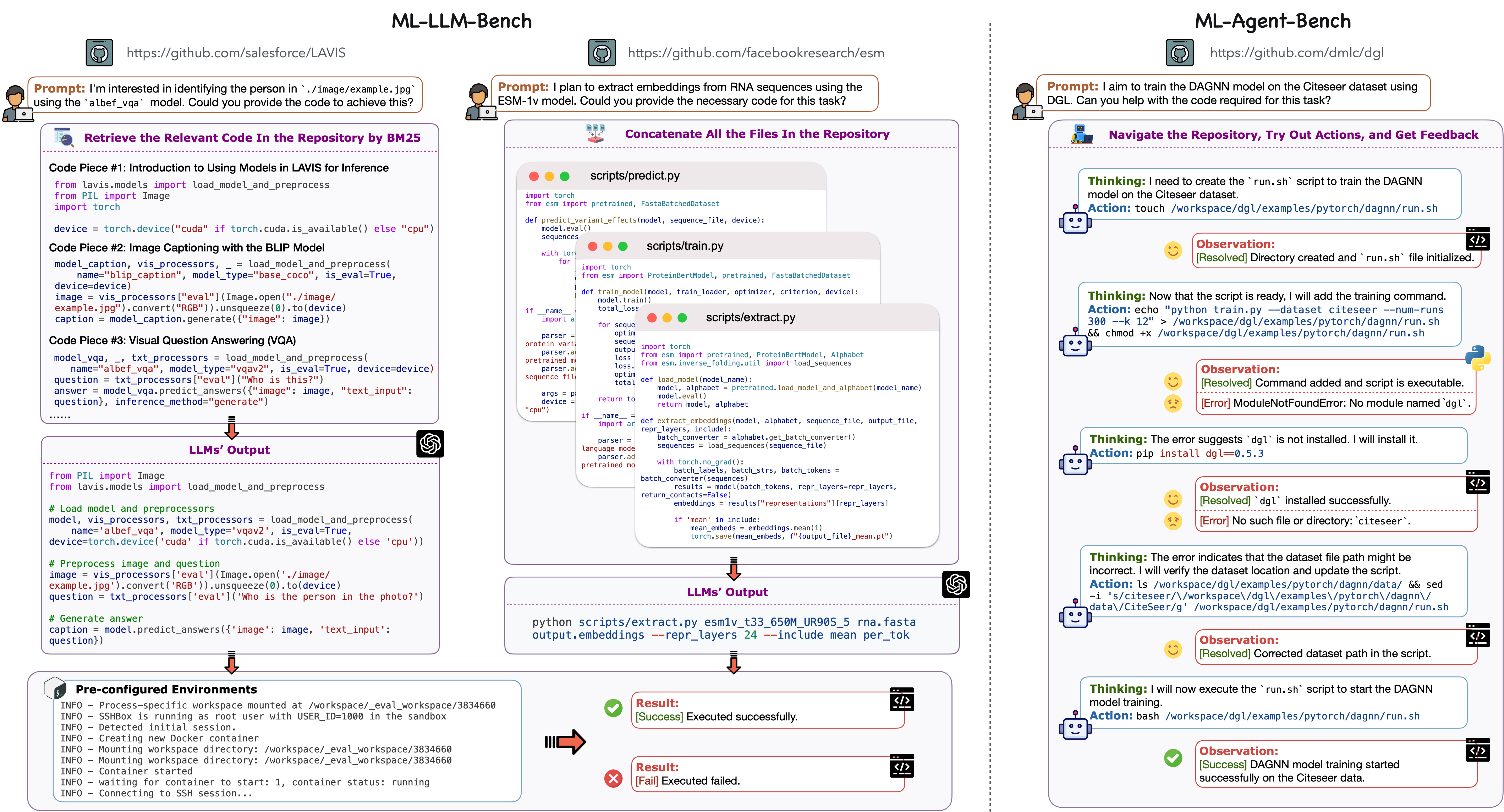}
\caption{\small The workflow of \textsc{ML-Bench}, including ML-LLM-Bench and ML-Agent-Bench. In ML-LLM-Bench, LLMs generate Python code or Bash scripts based on the prompt. The input to the LLMs could be code retrieved from a repository based on the prompt or a direct concatenation of all files. Their performance is evaluated within a pre-configured environment. Conversely, in ML-Agent-Bench, the agent must autonomously set up the environment and download necessary datasets to accomplish the task.
}
\label{fig:workflow}
\vspace{-.2cm}
\end{figure}

To better assess the abilities of LLMs versus autonomous AI agents, we present two distinct testing setups: \textsc{ML-LLM-Bench} and \textsc{ML-Agent-Bench}, as shown in Figure \ref{fig:workflow}.
(1) For LLMs, we evaluate the model's capacity to complete the task of calling code within our \textit{configured deployment environment}.
It benchmarks their competence in \textit{translating text instructions to simple bash code with clearly defined parameters}.
(2) However, a more arduous challenge lies in the end-to-end execution of tasks, starting from scratch. This involves initiating the code environment for a specific repository, where common pitfalls of environment setup, such as missing datasets or uninstalled packages, might occur. To evaluate agents in such a setup, we introduce a secure Linux sandbox environment, where agents can \textit{iteratively execute commands and code blocks to obtain feedback}.
The agent's actions involve multiple attempts, from reading files and understanding the repository to installing dependencies, preparing datasets, and finally writing bash code that calls the repository, thus emulating the full workflow of a human coder.

\vspace{-.3cm}

\begin{itemize}
    \item  ML-LLM-Bench: This benchmark component includes tasks that require generating both bash scripts and Python code. The diversity in task types reflects the varied nature of ML workflows, where both scripting and programming play essential roles.
    \item ML-Agent-Bench: In this more complex setup, agents must interact with the environment using a combination of bash commands and Python code with tools such as Jupyter Notebooks. This approach closely mimics ML practitioners' workflow, who often switch between command-line operations and code execution in interactive environments.
\end{itemize}

In our evaluation experiment on \textsc{ML-LLM-Bench}, we observe that GPT-4o outperforms other LLMs, being the sole model to surpass the \textbf{50\%} threshold in the Pass@5 metric (success rate within five tries).
It is noteworthy that in the same test set, our annotators‘ performance—computer science graduate students—stood at a success rate of \textbf{86.76\%}, with \textbf{59} out of \textbf{68} examples correctly executed. 
This indicates substantial room for improvement in current LLMs. However, the models did show performance improvements following instruction tuning on the training data (\textbf{8.85}$\rightarrow$\textbf{15.76} for CodeLlama).
Error analysis reveals that LLMs tend to generate hallucinations, predominantly producing incorrect parameters or referencing non-existent files. Generating bash scripts proved more challenging than generating Python code, pinpointing a capability bottleneck in LLMs. A critical insight from our study is the urgent need for LLMs to comprehend the long code context (the average length is around 150k tokens for the whole repository), not merely to generate code.
On the more challenging \textsc{ML-Agent-Bench} setup, GPT-4o scores \textbf{76.47\%} within the OpenDevin agent environment, where agents must configure their environment, navigate code repositories, and effectively generate the necessary code. 
Similar to Reflexion \citep{Shinn2023ReflexionAA} and CodeAct \citep{wang2024executable}, agents iteratively execute actions, refine their approach via feedback, and perform effectively in solving complex tasks. 
This underscores the potential of self-improvement and incorporating feedback from experience as alternatives to relying solely on instruction tuning with history training data to enhance LLM performance.

To sum up, while recent efforts have explored LLM-based agents for navigating GitHub repositories, such as conducting ML experiments in simplified environments \citep{huang2024mlagentbench} or resolving repository issues \citep{jimenezswe} (see Table \ref{tab:compare1}), ML-Bench addresses a distinct and critical challenge faced by many researchers: both \textit{setting up} and \textit{executing experiments} using research repositories in-the-wild, where code is often not well-maintained, and experiment settings are not easily configurable, such as:

\vspace{-.2cm}

\begin{itemize}

    \item  Unlike RepoBench \citep{Liu2023RepoBenchBR}, which focuses on repo-level code auto-completion, ML-Bench requires agents to generate and execute code to run experiments using research repositories. This involves more complex tasks, including environment setup, data preparation, and experiment execution. In our experiment, the models often struggle with this stage (the performance of the strongest model GPT-4o is 32.99\%).

    \item SWE-Bench \citep{jimenezswe} tasks agents with locating and modifying specific functions within a file to resolve an issue, providing a pre-deployed testing environment. In contrast, ML-Bench challenges agents to independently configure environments and download necessary data, mimicking real-world research scenarios more closely.

    \item ML-Bench evaluates the entire workflow of setting up and running ML experiments, from initial repository exploration to result interpretation. This end-to-end approach differentiates it from more narrowly focused benchmarks.

\end{itemize}









%% file: files/2-dataset.tex
\vspace{-.2cm}
\section{\benchname Construction}
\label{sec:related}

\vspace{-.2cm}

\begin{table*}[t!]
\centering
\caption{\small Comparison of benchmarks for repository-level code analysis: this comparison focuses on several key attributes across various benchmarks: (1) \textit{Repository Understanding}—the ability to comprehend and navigate the overall structure, dependencies, and functionality of an entire code repository beyond individual files; (2) \textit{Documentation Understanding}—the capability to interpret and utilize documentation elements such as README files to gain insights within the repository; (3) \textit{Cross-File Retrieval}-identifying relevant information across multiple files to complete tasks or resolve issues; (4) \textit{Package Installation}—installing dependencies required for the repository; (5) \textit{Data Downloading}—downloading data required for the task; and (6) \textit{Evalution}-the methods used to assess and measure the task performance.}

\resizebox{\linewidth}{!}{
\begin{tabular}{lccccc}
\toprule
\textbf{Criteria} & \textbf{\textsc{RepoEval} ~\citep{Zhang2023RepoCoderRC}} & \textbf{\textsc{RepoBench}~ \citep{Liu2023RepoBenchBR}} & \textbf{\textsc{MLAgentBench}~\citep{huang2024mlagentbench}} & \textbf{\textsc{SWE-Bench}~\citep{jimenezswe}} & \textbf{\textsc{ML-Bench (Ours)}} \\
\midrule
Repo. Understanding & \yes & \yes & \no  & \yes & \yes \\
Doc. Understanding  & \no  & \no  & \yes  & \no  & \yes \\
Cross-File Retrieval  & \no  & \yes  & \yes  & \yes & \yes \\
Package Installation       & \no  & \no  & \no  & \no  & \yes \\
Data Downloading      & \no  & \no  & \no  & \no  & \yes \\
Evaluation  & Similarity / Pass@K & Similarity & Test Accuracy & Success Rate & Pass@K / Success Rate \\
\# of Repositories & 14  & 3,116 & \slash  & 12   & 18 \\
\# of Tasks & 13,710 & 49,684 & 13 & 2,300 & 9,641 \\
\bottomrule
\end{tabular}
}
\label{tab:compare1}
\vspace{-.5cm}
\end{table*}

\subsection{Task Formulation and Design Principle}
\label{ssec:task_formulation}
\vspace{-.2cm}

\textsc{ML-Bench} aims to test LLM performance in real coding scenarios, which includes interpreting repository-level code and writing code to employ existing repositories to meet user demands. In our benchmark, we consider a scenario where a language model, given a GitHub repository $\mathcal{F}$, accesses all files $f \in \mathcal{F}$. Upon receiving an instruction $i$ with arguments $\mathrm{arg}_i \in \mathcal{A}$, the model must generate executable code $c$ that utilizes functions or models from the repository in line with the user instruction $i$ and arguments $\mathrm{arg}_i$.

For \textsc{ML-Bench} settings, (1) \textsc{ML-LLM-Bench} provides a complete build environment, allowing us to test output bash scripts generated by LLMs within a Docker environment regarding the correctness and excitability. 
(2) \textsc{ML-Agent-Bench} provides agents with access to an empty Docker environment without essential packages. Agents must attempt to download the requirements for each user instruction involving the installation of new datasets or Python packages themselves. 
This design ensures that our testing framework aligns with practical application workflow.

In addition, to address potential data contamination, despite the possibility that code in GitHub repositories may have been exposed to LLMs during pretraining, we mitigate the risk by manually rewriting inputs and outputs and verifying our dataset's uniqueness against internet searches, ensuring our outputs are free from such contamination.

\begin{table*}[!hbt]
\vspace{-.1cm}
\centering
\caption{Detailed breakdown of the number of bash script and Python code samples for each repository. The test set contains samples from \textbf{14} repositories, while the train set includes \textbf{4} additional repositories for the OOD setting. A quarter subset of the test set is also shown.
All repository names are hyperlinked for direct access to the corresponding GitHub.
}
\small
\resizebox{0.7\linewidth}{!}{
\begin{tabular}{llrrrrrr}
\toprule
\multicolumn{2}{c}{\multirow{2}{*}{\textbf{Repository}}} & \multicolumn{2}{c}{\textbf{Train Set}} & \multicolumn{2}{c}{\textbf{Test Set}} & \multicolumn{2}{c}{\textbf{1/4 Test Set}} \\
\cmidrule(lr){3-4} \cmidrule(lr){5-6} \cmidrule(lr){7-8}
\multicolumn{2}{c}{} & \textbf{Scripts} & \textbf{Code} & \textbf{Scripts} & \textbf{Code} & \textbf{Scripts} & \textbf{Code} \\
\midrule
\rowcolor[RGB]{238, 255, 255}\multicolumn{8}{c}{\textbf{In-Distribution (ID)}} \\
\multicolumn{2}{l}{\href{https://github.com/NVIDIA/vid2vid}{Video-to-Video (vid2vid)}} & 46 & 0 & 13 & 0 & 4 & 0 \\
\multicolumn{2}{l}{\href{https://github.com/deep-floyd/if}{IF by DeepFloyd Lab (If)}} & 168 & 175 & 10 & 11 & 4 & 2 \\
\multicolumn{2}{l}{\href{https://github.com/dmlc/dgl}{Deep Graph Library (DGL)}} & 553 & 0 & 21 & 0 & 5 & 0 \\
\multicolumn{2}{l}{\href{https://github.com/eriklindernoren/PyTorch-GAN}{Pytorch-GAN (Py-GAN)}} & 1080 & 0 & 30 & 0 & 8 & 0 \\
\multicolumn{2}{l}{\href{https://github.com/facebookresearch/esm}{ESM}} & 563 & 58 & 15 & 2 & 4 & 1 \\
\multicolumn{2}{l}{\href{https://github.com/google-research/bert}{BERT}} & 962 & 0 & 22 & 0 & 6 & 0 \\
\multicolumn{2}{l}{\href{https://github.com/mlfoundations/open_clip}{OpenCLIP}} & 646 & 691 & 10  & 1 & 3 & 0 \\
\multicolumn{2}{l}{\href{https://github.com/salesforce/lavis}{Lavis - A Library for Language (Lavis)}} & 76 & 205 & 4 & 23 & 1 & 6 \\
\multicolumn{2}{l}{\href{https://github.com/thuml/Time-Series-Library}{Time Series Library (TSL)}} & 1449 & 0 & 14 & 0 & 4 & 0 \\
\multicolumn{2}{l}{\href{https://github.com/xmu-xiaoma666/External-Attention-pytorch}{Attention Series (EAP)}} & 95 & 5 & 24 & 0 & 5 & 0 \\
\midrule
\rowcolor[RGB]{238, 255, 255}\multicolumn{8}{c}{\textbf{Out-Of-Distribution (OOD)}} \\
\multicolumn{2}{l}{\href{https://github.com/IDEA-Research/Grounded-Segment-Anything}{Grounded-SAM}} & \slash & \slash & 12 & 8 & 2 & 3 \\
\multicolumn{2}{l}{\href{https://github.com/huggingface/pytorch-image-models}{PyTorch Image
Models (Py-IM)}} & \slash & \slash & 5 & 0 & 1 & 0 \\
\multicolumn{2}{l}{\href{https://github.com/microsoft/muzic}{muzic}} & \slash & \slash & 17 & 1 & 4 & 1 \\
\multicolumn{2}{l}{\href{https://github.com/vinits5/learning3d}{Learning3D}} & \slash & \slash & 17 & 0 & 4 & 0 \\
\multicolumn{2}{l}{\href{https://github.com/Stability-AI/stablediffusion}{Stable Diffusion (SD)}} & 2253 & 0 & \slash & \slash & \slash & \slash \\
\multicolumn{2}{l}{\href{https://github.com/black0017/MedicalZooPytorch}{Medical Zoo (MedZooPy)}} & 490 & 0 & \slash & \slash & \slash & \slash \\
\multicolumn{2}{l}{\href{https://github.com/brightmart/text_classification}{Time Series Library (TCL)}} & 196 & 0 & \slash & \slash & \slash & \slash \\
\multicolumn{2}{l}{\href{https://github.com/tensorflow/tensor2tensor}{Tensor2Tensor}} & 0 & 248 & \slash & \slash & \slash & \slash \\
\midrule
\multicolumn{2}{l}{Total} & \textbf{8577} & \textbf{736} & \textbf{214} & \textbf{46} & \textbf{55} & \textbf{13} \\
\bottomrule
\end{tabular}
}
\label{tab:num-types}
\vspace{-.3cm}
\end{table*}

\subsection{Summary of Data}
\vspace{-.2cm}

We construct \textsc{ML-Bench} from 18 diverse repositories, each reflecting varying complexity and tasks, while filtering out substandard samples. The data quantities and breakdown per repository are detailed in Table~\ref{tab:num-types}. Regarding the code language, our annotated output includes both \textbf{bash scripts}, which invoke Python files with specific arguments, and \textbf{Python code}, which calls functions directly from the repository. Bash scripts significantly outnumbered Python code snippets.

On average, each repository contributed approximately 480 examples, summing up to 9,641 examples.
For our experiments involving the fine-tuning of open-source LLMs, we split the dataset based on code origin: The \textbf{In-Distribution (ID)} approach utilizes data from the same repository both for training and testing, allowing repository-specific code to be exposed to models during fine-tuning. In contrast, the \textbf{Out-Of-Distribution (OOD)} method employs disjoint sets for training and testing, encompassing eight repositories—half for model training and the remaining for evaluation. The overall statistics and further detailed data metrics for each repository utilized can be found in Appendix~\ref{appendix:details_of_selected_github_repositries}.

\vspace{-.2cm}

\subsection{Data Collection and Annotation Pipeline}
\label{ssec:annotation_pipeline}
\vspace{-.2cm}

\begin{figure*}[t]
\centering
\vspace{-.2cm}
\includegraphics[width=\textwidth]{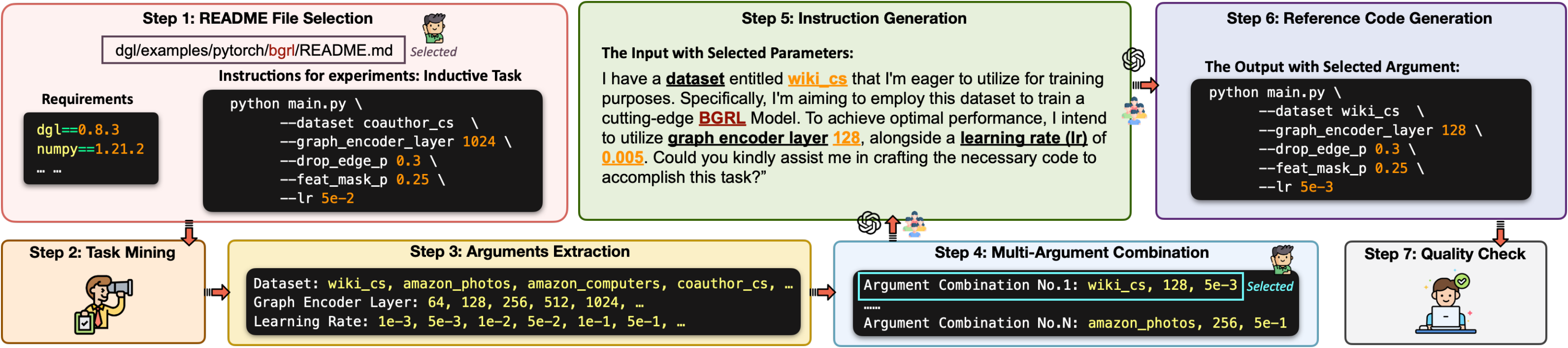}
\caption{The construction pipeline of our \benchname.}
\label{fig:data construction}
\vspace{-.4cm}
\end{figure*}

Eight computer science graduate students with proficient programming abilities contributed to the data annotation of \textsc{ML-Bench}, with each repository's related data being the responsibility of one annotator and an additional reviewer ensuring data accuracy. Annotators were permitted to use GPT-4 to expedite the annotation process, although manual verification and adjustments were required. Annotating a repository took approximately 5-10 hours. The annotation workflow is shown in Figure ~\ref{fig:data construction}:

\textbf{(1) README file Selection}: Annotators commenced by meticulously reviewing repository contents to identify all README files, including those within various subdirectories, each covering different functionalities. On average, a GitHub repository included 12 README pages, with one notable repository, DGL, comprising 154 README files.
\textbf{(2) Task Mining}: Annotators identify practical tasks from README files, along with corresponding code examples, averaging nine tasks per repository, thus capturing the representative functionality of each GitHub repository.
\textbf{(3) Arguments Extraction}: Beyond task identification, annotators with machine learning expertise extracted key parameters essential for task completion, targeting representative parameters commonly employed in practical experiments.
\textbf{(4) Multi-Argument Combination}: With tasks and arguments identified, annotators create diverse combinations of argument values, essential for constructing scenarios that represent real-world applications of repository code.
\textbf{(5) Instruction Generation}: Utilizing ChatGPT, we generate task-specific instructions for each argument set, applying templates to ensure diversity and explicit argument inclusion, detailed in Appendix ~\ref{appendix:templates_diverse_instruction_construction}.
\textbf{(6) Reference Code Generation}: For each argument combination, we develop code templates to generate accurate ground truth code for the targeted tasks.
\textbf{(7) Quality Check}: The dataset underwent stringent quality checks, particularly for code executability and argument accuracy, with any non-compliant data being revised or discarded. This ensures that the instructions precisely align with user requirements, thereby upholding the integrity and applicability of the \textsc{ML-Bench} benchmark.

%% file: files/4-setup.tex
\vspace{-.2cm}

\section{\textsc{ML-LLM-Bench} Experiments}
\label{sec:setup}

\vspace{-.2cm}

\subsection{\textsc{ML-LLM-Bench} Setup}
\label{ssec:experimental_setup}
\vspace{-.2cm}

Our experimental inputs include human instructions and the entire repository code (including README files). We present three distinct experimental setups to evaluate the models. Given that current models cannot process the entire code context, the three scenarios range from ideal to extreme.
\textbf{Oracle Segment (Oracle):} For the Oracle setup, annotators identify and record crucial segments within README files — referred to as "Oracle Segments" — that contain necessary codes and textual explanations pivotal for completing the prescribed tasks. These segments serve as the foundational source to derive the ground truth code, ensuring that models can access all critical evidence when generating code. \textbf{BM25 Retrieval (Retrieval):} In this setup, we employ a BM25 retriever to extract segments from the repository's documentation, including README files, that are relevant to the given instructions. This method aims to mimic a more automated way of narrowing down necessary information without human pre-selection. \textbf{Code (Code):} This setting exposes the model to the entire code repository. All files within the repository, with README files placed at the forefront, are presented as input to the model. Due to model context limitations, texts are truncated when necessary, and potential information loss is analyzed and documented in Appendix~\ref{appendix:details_of_selected_github_repositries}. Please refer to the supplementary materials for further details on implementing the BM25 retriever.


\begin{table*}[!]
\centering
\caption{\footnotesize Pass@1/5 scores for models on the \textsc{Scripts|Code} (bash script, Python code) partition of \textsc{ML-LLM-Bench}. 
\dag denotes instruction-tuned models. Results are shown for the \textbf{Oracle}, \textbf{Code}, and \textbf{Retrieval} settings. Results under ID and out-of-distribution (OOD) are reported after instruction fine-tuning.}
\resizebox{\linewidth}{!}{
\begin{tabular}{l l c c c c c c}
\toprule
\multicolumn{2}{c}{\multirow{2}{*}{\textbf{Models}}} & \multicolumn{2}{c}{\textbf{Oracle}$^{\textsc{Scripts|Code}}$} & \multicolumn{2}{c}{\textbf{Code}$^{\textsc{Scripts|Code}}$} & \multicolumn{2}{c}{\textbf{Retrieval}$^{\textsc{Scripts|Code}}$} \\
\cmidrule(lr){3-4} \cmidrule(lr){5-6} \cmidrule(lr){7-8}
\multicolumn{2}{c}{} & \textbf{Pass@1} & \textbf{Pass@5} & \textbf{Pass@1} & \textbf{Pass@5} & \textbf{Pass@1} & \textbf{Pass@5} \\
\midrule
\multicolumn{2}{l}{Human} &  \slash & \slash &$86.76$ & \slash & \slash & \slash  \\
\rowcolor[RGB]{234, 238, 234} \multicolumn{8}{c}{\it Close-Source LLMs 
} \\
\multicolumn{2}{l}{GPT-4o} & $36.42^{31.37|56.83}$ & $50.13^{44.26|78.89}$ & $32.99^{31.44|39.87}$ & $46.20^{43.58|61.54}$ & $26.16^{19.47|55.52}$ & $30.44^{24.73|76.92}$ \\
\multicolumn{2}{l}{GPT-4} &  $33.82^{29.09|53.85}$ & $48.53^{41.81|76.92}$ & $30.88^{29.09|38.46}$ & $45.59^{41.82|61.54}$ & $22.06^{14.55|53.85}$ & $27.94^{16.36|76.92}$ \\
\multicolumn{2}{l}{GPT-3.5} & $27.94^{21.81|53.85}$ & $38.23^{30.91|69.23}$ & $15.07^{0.09|38.46}$ & $30.14^{23.64|53.85}$ & $13.70^{5.45|46.15}$ & $24.66^{14.55|69.23}$ \\
\multicolumn{2}{l}{Claude-3-Opus} & $25.52^{12.15|67.39}$ & $36.92^{27.57|80.43}$ & $13.46^{0.70|43.48}$ & $35.39^{30.37|58.70}$ & $10.00^{3.27|41.30}$ & $22.69^{11.22|76.09}$ \\
\multicolumn{2}{l}{Claude-3-Sonnet} & $21.92^{18.18|38.46}$ & $34.25^{27.27|61.54}$ & $27.40^{25.45|30.76}$ & $35.62^{30.91|53.85}$ & $9.59^{3.64|38.46}$ & $20.55^{9.09|69.23}$ \\
\multicolumn{2}{l}{Claude-3-Haiku} & $18.46^{11.68|50.00}$ & $30.38^{20.09|78.26}$ & $25.38^{22.90|36.96}$ & $32.31^{28.04|52.17}$ & $8.08^{3.74|28.26}$ & $16.92^{7.48|60.87}$ \\

\midrule
\rowcolor[RGB]{234, 238, 234} \multicolumn{8}{c}{\it Open-Source LLMs  
} \\

\multicolumn{2}{l}{CodeLlama-7b} & $8.85^{3.37|32.60}$ & $21.15^{11.68|65.22}$ & $1.54^{0.47|6.52}$ & $8.85^{2.80|36.96}$ & $0.77^{0.00|4.34}$ & $8.85^{2.80|36.96}$ 
\\




\multicolumn{2}{l}{DeepseekCoder-6.7b} & $9.23^{0.46|30.43}$ & $24.23^{14.02|71.74}$ & $3.85^{1.89|13.04}$ & $10.38^{6.07|30.43}$ & $5.00^{3.27|13.04}$ & $14.23^{9.81|34.78}$ \\


\multicolumn{2}{l}{LLaMA-2-7b} & $2.27^{0.13|5.70}$ & $4.77^{2.47|6.22}$ & $0.00$ & $0.00$ & $0.00$ & $0.00$  \\

\midrule
\rowcolor[RGB]{234, 238, 234} \multicolumn{8}{c}{\it Finetuned LLMs w/ the Out-Of-Distribution (OOD)} \\

\multicolumn{2}{l}{CodeLlama-7b \dag} & $15.76^{12.14|32.61}$ & $28.46^{19.62|69.57}$ & \slash & \slash & $1.92^{0.47|8.70}$ & $5.38^{1.40|23.91}$ \\

\multicolumn{2}{l}{DeepseekCoder-6.7b \dag} & $16.15^{14.95|34.78}$ & $31.15^{24.30|58.70}$ & \slash & \slash & $10.38^{6.54|28.26}$ & $26.15^{17.29|67.39}$ \\

\multicolumn{2}{l}{LLaMA-2-7b \dag} & $5.31^{2.47|10.86}$ & $6.03^{3.12|11.64}$ & \slash & \slash & $2.77^{1.30|5.34}$ & $5.31^{2.47|10.86}$ \\

\midrule
\rowcolor[RGB]{234, 238, 234} \multicolumn{8}{c}{\it Finetuned LLMs w/ the In-Distribution (ID)} \\
\multicolumn{2}{l}{CodeLlama-7b \dag} & $17.69^{15.42|28.26}$ & $30.77^{21.96|71.74}$ & \slash & \slash & $2.69^{0.47|13.04}$ & $9.62^{3.27|39.13}$ \\
\multicolumn{2}{l}{DeepseekCoder-6.7b $\dag$} & $21.92^{12.16|65.22}$ & $30.77^{20.56|78.26} $& \slash & \slash & $2.69^{1.40|8.70}$ & $10.00^{5.61|30.43}$\\
\multicolumn{2}{l}{LLaMA-2-7b \dag} & $6.54^{2.33|26.09}$ & $8.38^{4.45|32.17}$ & \slash & \slash & $1.15^{0.00|6.52}$ & $3.08^{4.67|15.22}$ \\
\bottomrule
\end{tabular}
}
\label{tab:main-ppass}
\end{table*}

\begin{table*}[!ht]
\centering
\caption{Agent evaluation results on the \textsc{ML-Agent-Bench}. The success rate, number of solved instances, and the average cost per solved instance are reported for each agent and language model combination. $\dag$ Evaluation is conducted on a quarter subset of the test set due to budget constraints. $\S$ OpenDevin is available on https://github.com/OpenDevin/OpenDevin.
}\vspace{-.2cm}

\vspace{3pt}
\resizebox{\linewidth}{!}{
\begin{threeparttable}
\begin{tabular}{l|l|rrr}
\toprule
\textbf{Agent}
& \textbf{Model Name}
& \textbf{Success Rate$^\dag$} (\%)
& \textbf{\# of Solved Instances} 
& \textbf{\$ Avg. Cost} \\
\toprule
AutoGen \citep{wu2023autogen} & \texttt{gpt-4-1106-preview} & $8.82$ &  $6$ & $1.28$ \\
SWE-Agent \citep{yang2024swe} & \texttt{gpt-4-1106-preview} & $42.64$ &  $29$ & $1.91$ \\
Aider \citep{aider} & \texttt{gpt-4o} & $64.38$ &  $47$ & - \\
\midrule
\multirow{3}{*}{OpenDevin$^\S$} 
& \texttt{gpt-4o-2024-05-13} & $76.47$ & $51$ & $0.25$ \\
& \texttt{gpt-4-1106-preview} & $58.82$ & $40$ & $1.22$ \\
& \texttt{gpt-3.5-turbo-16k-0613} & $13.23$ & $9$  & $0.12$\\
\bottomrule
\end{tabular}
\end{threeparttable}
}
\label{tab:swe_bench}
\vspace{-.2cm}
\vspace{-.2cm}
\end{table*}

\vspace{-.2cm}

\subsection{Evaluation Metrics}
\label{ssec:evaluation_metrics}

\vspace{-.2cm}
The generated code must be executable and adhere to the parameters outlined in the user instructions. We use Pass@K as our metric for evaluation, with K representing the number of generation attempts allowed. Pass@K measures the likelihood of the model producing at least one correct code execution in those K tries (given unit tests).
\vspace{-.2cm}
\subsection{Experimental Results}

\vspace{-.2cm}As presented in Table \ref{tab:main-ppass}, we conducted evaluations on a set of LLMs including GPT-4o (model name: gpt-4o-2024-05-13), GPT-4 (model name: gpt-4-1106-preview), GPT-3.5 (model name: gpt-3.5-turbo-16k-0613), and the Claude 3 model family (Claude-3-Opus, Claude-3-Sonnet, Claude-3-Haiku). Moreover, we selected CodeLlama-7b-Instruct, DeepSeek-Coder-6.7b-Instruct, and Llama-2-7b-chat-hf (LLaMA 2) to explore the effects of fine-tuning with an 8k token length limit with 4 A100s.
The findings suggest that while GPT-4o exhibited the highest scores across the test cases, the untrained models, such as LLama-2-7b, performed poorly on the \textsc{ML-LLM-Bench}, even after in-distribution (ID) fine-tuning. Fine-tuning on out-of-distribution (OOD) data indicated that models could benefit from training on similar tasks, though not necessarily from the same repository. Moreover, the performances on ID data implied that even after task-relevant fine-tuning, the results from 7B-scale open-source models could not outperform the closed-source counterparts.
The oracle setting outcomes demonstrate that providing models with the correct reference solutions is effective for task completion. A retrieval approach not specifically designed for the task might lead to suboptimal results, potentially hindering performance.

\section{\textsc{ML-Agent-Bench} Experiments}



\subsection{\textsc{ML-Agent-Bench} Setup}
\begin{figure}[!h]
\centering
\vspace{-.16cm}
\includegraphics[width=0.95\textwidth]{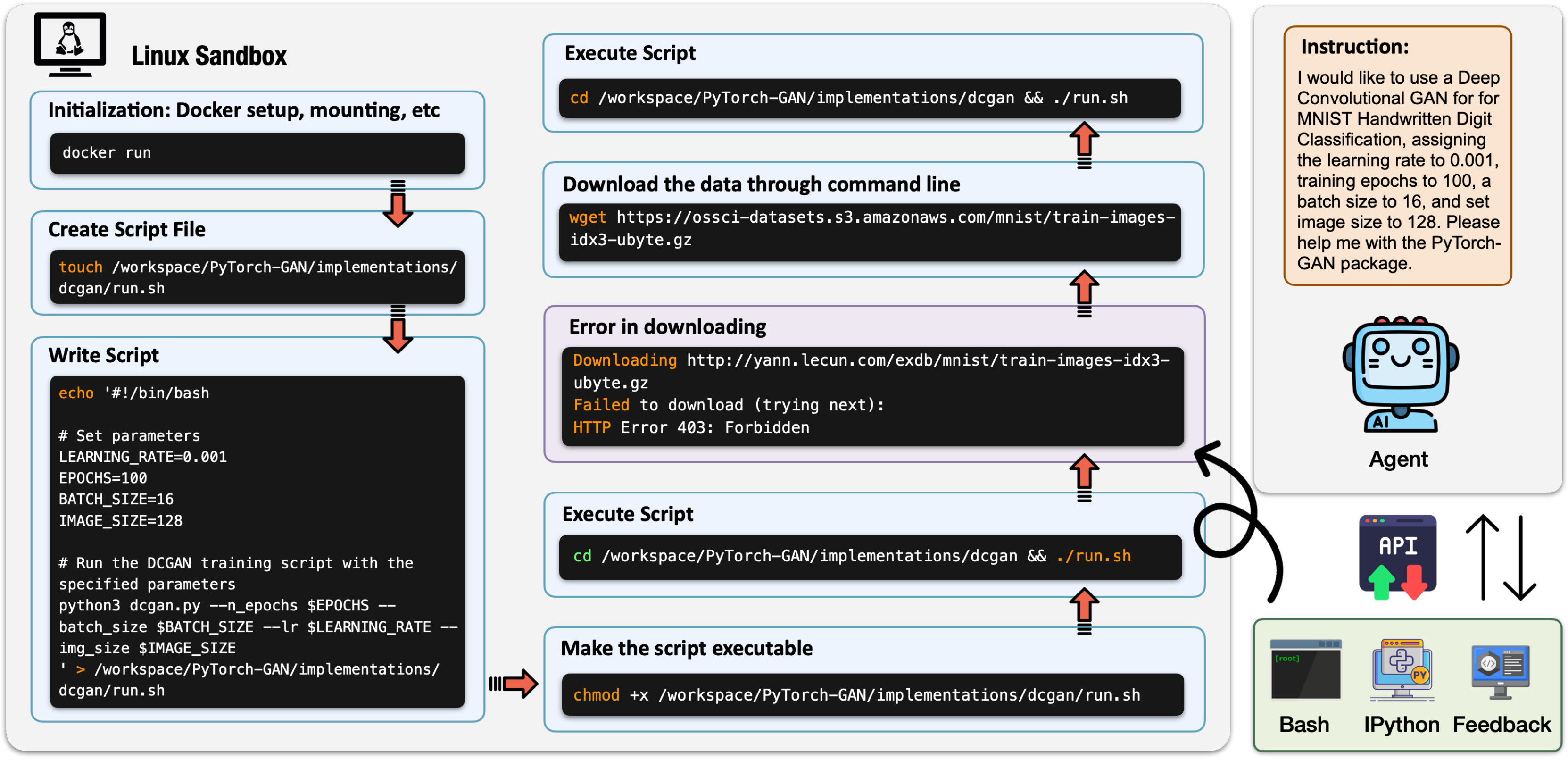}
\caption{An example of the \textsc{ML-Agent-Bench} evaluation runtime}
\label{fig:compare_data}
\vspace{-.2cm}
\end{figure}

In \textsc{ML-Agent-Bench}, as shown in Figure~\ref{fig:compare_data}, we provision a sandbox environment as the testing ground for agents. The sandbox offers a fundamental setup, such as a configurable Docker image, allowing agents to modify and execute commands freely within the simulation. Agents are granted the ability to execute bash scripts or interact with IPython notebooks.
The agents must interact with this environment, perusing code within repositories—regardless of the extended code or required parameters—to accumulate comprehensive information. This process necessitates successive actions, with the agent autonomously determining the correctness of each step and iteratively refining its approach upon encountering errors. We expect the agents' outputs to differ from previous LLM settings due to the dynamic and interactive nature of the tasks.

Taking cues from projects such as CodeAct \citep{wang2024executable}, recent agent frameworks, including SWE-Agent \citep{yang2024swe}, Aider \citep{aider}, OpenDevin (https://github.com/OpenDevin/OpenDevin), provide a well-defined suite of impactful actions that bridge the agent with its operational environment. These agents are designed to mimic the workflow of human programmers, thoroughly parsing and employing a repository. To facilitate this, the agents are endowed with the ability to execute \textit{any} Python code and bash commands within a secure and isolated Linux OS sandbox, providing an ideal setting for our benchmark evaluations.
In each instance, \textsc{ML-Agent-Bench} initiates an isolated docker container sandbox where all agents' bash commands are executed, with the outcomes returned as observations.
Different agent frameworks implement environmental interactions in varying ways, with each action yielding observations for AI agents. Here \textsc{ML-Agent-Bench} is essentially assessing the effectiveness of different environments. 
In \textsc{ML-Agent-Bench}, a configurable workspace directory contains repositories agents are to handle, installed within this safe sandbox, providing access for the agents to interact with and process as needed.
For evaluation, instead of using the Pass@K metric in the \textsc{ML-LLM-Bench} setting, we are emphasizing the agent's effectiveness in fulfilling user requirements through interactive execution rather than predetermined outputs (\textbf{Success Rate}). Success is characterized not just by code generation but also by the agent's ability to alter or create new files, such as a \textit{run.sh}, to achieve the execution required by the instruction.

\vspace{-.2cm}
\subsection{Experimental Results}
In Table \ref{tab:swe_bench}, we detail the performance of various agents such as AutoGen, SWE-Agent, and Aider, as well as OpenDevin equipped with diverse GPT language models, evaluated on a quarter subset of the test set. OpenDevin, utilizing GPT-4o (model name: gpt-4o-2024-05-13), achieved the best results, striking an excellent balance between cost and performance. The success rate, the number of instances successfully solved, and the average cost per solved instance were the critical metrics for this evaluation. As demonstrated by the varied performance of agents using the identical \textit{gpt-4-1106} model, it becomes evident that the choice of agent framework significantly impacts the effectiveness of an agent. This discrepancy in success rates and average costs accentuates the potential for future advancements in agent architecture to enhance performance further.

%% file: files/5-experiments.tex
\vspace{-.2cm}

\section{Analysis}
\label{sec:analysis}

\vspace{-.2cm}

\subsection{Data Leakage} 
\vspace{-.2cm}

Our early experiments found that models tended to generate Python code instead of bash script regardless of whether the input is script or code. This is related to the fact that model trainers expect they can get high scores on other code benchmarks, causing these polished repositories to be more likely to be put into the training set compared to lesser-known ones, and code data is a larger percentage than script data.

To mitigate the influence of data leakage, the type and parameters of the generated result are checked to see whether they are the same as those present in the provided documents and user instructions before execution. We show the updating status for all repositories and the cutoff date of training data for models in Appendix~\ref{appendix:details_of_selected_github_repositries}.




\begin{figure*}[htbp]
\vspace{-.4cm}
\centering
\includegraphics[width=0.95\textwidth]{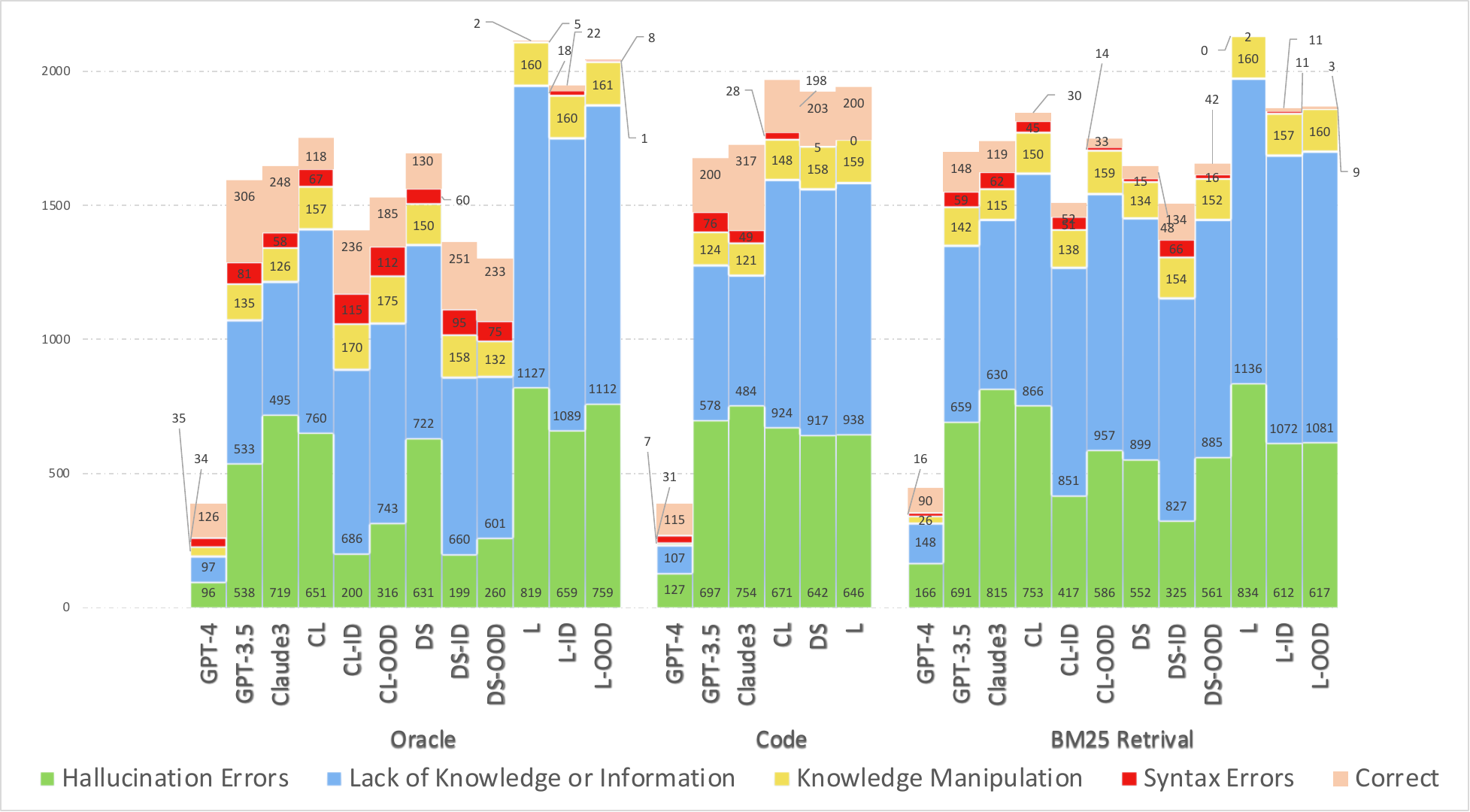}
\caption{Quantification of models and settings errors with five attempts. The total statistic results are 1300 for the full test set. Statistical results that exceed these numbers are caused by multiple errors made on one result simultaneously. For models, CL means CodeLlama, DS means deepseek-coder, and L means LLaMA.
Raw means that the model is not fine-tuned. ID means that the model is fine-tuned in an in-distribution setting. OOD means that the models are fine-tuned in an out-of-distribution setting. Here, Claude3 means Claude-3-Haiku.
}
\label{fig:model_setting_analysis}
\end{figure*}


\vspace{-.2cm}

\subsection{Error Analysis}

By analyzing the execution log, we find that the errors for models and agents in \benchname fall into five categories:

\textbf{Hallucination Errors (E1)}: 
These errors include instances when the models misinterpreted the user's intention, misplaced Python code and bash script, or generated random or irrelevant code.

\textbf{Lack of Knowledge or Information (E2)}: 
This type of error primarily stems from the model's inability to fulfill user requirements based on crucial parameters. Possible types of errors are as follows:
\vspace{-0.2cm}

\begin{enumerate}
    \item Code inner information.
The models sometimes lack sufficient information necessary to satisfy the user's requirements. For instance, this deficiency might involve missing parameter names (\texttt{----lr} and \texttt{----learning--rate}) or unavailable options (it only accepts 'Citeseer' when the input given was 'citeseer').

    \item Domain knowledge.
The models sometimes lack the domain-specific knowledge required to handle certain instances. For example, in BERT, the models simultaneously generated \texttt{----case=True} and \texttt{----model=uncased}.

    \item Grammar knowledge.
This happens when the models incorrectly identify and handle certain command line symbols. 
Like the \$ symbol, which could affect execution.

    \item Local data information.
The models were not supplied with enough crucial parameter information for specific commands, leading to the inability to find paths and successful execution. 
While less common, this error was particularly notable in the OpenCLIP. 
\end{enumerate}

\textbf{Knowledge Manipulation (E3)}:
Take BERT, for instance, where the model needed to integrate \texttt{DIR=/model/} and \texttt{----model\_dir=\$DIR} to form \texttt{----model\_dir=/model}. 
There were also cases where it couldn't merge \texttt{/model\_name in /data} into a proper path format like \texttt{/data/model\_name}. 
Similar incidents were observed in OpenCLIP.

\textbf{Syntax Errors (E4)}: These errors cover instances of incorrect code generation of syntax errors instead of hallucination, mainly Python syntax errors such as the use of undefined variables. These errors arise from cases that prefer generating Code.

\textbf{Operational Error (E5 -- \textsc{ML-Agent-Bench} Only)}: These errors only occur in \textsc{ML-Agent-Bench}. Less powerful agents, such as those using base models like GPT-3.5, may have difficulty understanding the task in the execution sandbox. In these cases, the agent gives up easily or exits without finishing the tasks. Interestingly, these errors are less observed in advanced model settings.

On \textsc{ML-LLM-Bench}, we find that Claude 3 has more hallucinations than GPTs for closed-source models. However, its ability to fulfill requirements (represented by E2) is better than that of GPT-3.5. 
Under the retrieval setting, neither GPT-3.5 nor Claude 3 exhibits an increase in hallucination but an increase in the proportion of E2 compared to the Code setting. For GPT-4, both E1 and E2 increased because GPT-4 generates code without considering the content in cases involving task-irrelevant information, except for GPT-3.5 and Claude 3.

Compared to the Code setting, Oracle provision decreases the quantities of E1 and E2, while the differences in E3 and E4 are insignificant. This suggests that whether an Oracle is provided does not significantly affect the closed-source models' knowledge manipulation and Python code generation abilities. We tend to attribute these to the models' inherent ability rather than the reference.



The \textsc{ML-Agent-Bench} showcases potential in incorporating feedback from the experience, leading to fewer E1 and E4 errors. Yet, compared to \textsc{ML-LLM-Bench}, \textsc{ML-Agent-Bench} is more prone to E2 and E3 errors due to file type discrepancies from the set output type.
Because of the great flexibility of the sandbox environment, we observe increased hallucinations with each step, including installing conflicting dependencies and navigating incorrect or even illegal directories.

Meanwhile, E5 happens in less powerful agents. The agent often ignores the instruction to export the final solution to \texttt{run.sh}, or refuses to answer the question immediately after entering the virtual sandbox.
Compared to GPT-4, GPT-4o shows enhanced abilities in sourcing relevant information from repositories, leading to markedly fewer errors associated with E2 and E3. Yet, it exhibits a tendency toward more frequent hallucinations than GPT-4, for instance, installing incorrect packages.

For a detailed error analysis and quantitative performance visualization on both setups, refer to Appendix~\ref{appendix:details_of_selected_github_repositries}.

%% file: files/6-conclusion.tex
\vspace{-0.3cm}
\section{Conclusion}
\label{section:conclusion}
\vspace{-0.3cm}



Our study evaluates the performance of Large Language Models (LLMs) like GPT-4 in repository-scale code understanding and autonomous code interaction.
Our evaluation includes two setups: \textsc{ML-LLM-Bench} for text-to-code conversion and \textsc{ML-Agent-Bench} for end-to-end task execution. Results show GPT-4 achieving a Pass@5 rate over 50\% in \textsc{ML-LLM-Bench} and a 76.47\% success rate in \textsc{ML-Agent-Bench}, highlighting areas for improvement in handling hallucinations and bash script generation.

%% file: files/8-ethics+repro.tex
\section*{Limitation}
\label{sec:limitation}

Our study, while comprehensive within its scope, is subject to certain limitations that stem primarily from linguistic and data source constraints.

\paragraph{Linguistic Limitation - English as a Working Language}

We exclusively focused on English for our analyses and model development. This choice, while pragmatic due to English's prevalence in scientific literature and technical documentation, inherently limits the generalizability of our findings. English, as a language, possesses unique syntactic and semantic structures that may not be representative of other languages. Consequently, the applicability of our results to non-English contexts is uncertain.
This linguistic limitation also restricts the diversity of perspectives and cultural nuances that non-English documents could offer. 

\paragraph{Data Source Limitation - Reliance on GitHub Repositories in English}

Our reliance on GitHub repositories with documents exclusively in English introduces a selection bias. GitHub, while rich in open-source projects and documentation, may not comprehensively represent the broader landscape of software development practices and trends globally.
This choice potentially overlooks significant contributions and insights from non-English-speaking communities. 
This limitation might impact the development of tools and models tailored to a more diverse set of programming environments and community needs. 


\paragraph{Methodological Limitation - Relying on Pre-built Machine Learning Packages}

In our methodology, we utilized existing machine learning packages instead of developing algorithms from scratch. While this approach allowed us to leverage well-established, tested, and optimized tools, it also introduces certain constraints.
Dependence on pre-built packages means our work is confined to the capabilities and limitations of these tools. This reliance could limit our ability to fully explore novel or unconventional approaches possible with custom-built algorithms.
Moreover, this choice potentially impacts the reproducibility and customization of our findings. Researchers who seek to build upon our work may encounter similar constraints imposed by the pre-built packages we utilize. 
These limitations can hinder innovation and adaptation in different contexts or for specific usage.

\paragraph{Scope Limitation - Tasks Limited to README File Descriptions}

By strictly adhering to the specified tasks, our study may overlook potential applications or challenges not explicitly documented in README. This limitation can result in a narrower understanding of the tools we examined, as it fails to explore their full potential and applicability.
The reliance on README descriptions also assumes that these documents comprehensively and accurately reflect all relevant aspects of the repositories, which may not always be accurate. Important tasks or nuances might be undocumented or underrepresented in these files.

\section*{Ethics Statement}
\label{sec:ethics_statement}

In our work, we have carefully considered the ethical implications of our work, particularly in data annotation and related activities. Our methodologies and processes have been meticulously designed to ensure they are free from moral concerns. We affirm that our research practices, including data handling, have been conducted with the utmost integrity and in compliance with ethical standards.

Our approach has been guided by principles prioritizing respect for data integrity, transparency in our methods, and adherence to established ethical guidelines. 



\newpage

%% file: files/9-appendix.tex
\newpage
\appendix

\section{Related work}
\label{sec:related_work}

\subsection{Code Generation}
\label{sec:related_work_main}
\vspace{-0.3cm}


Code generation in natural language processing (NLP) has been a significant research topic, leading to the development of various methodologies and benchmarks, as seen in  \citep{cassano2022multiple, chen2021codex, christopoulou2022pangucoder, Li2022, orlanski2023measuring, Tang2023BioCoderAB, tang2023struc, wang2023codet5plus}.
Current benchmarks primarily aim to enhance function-level code generation capabilities. However, \benchname diverges by integrating code generation to streamline the usage of repositories within real-world workflows. For a comparative overview, see Table ~\ref{tab:code_bench}.
The goal of function-level code generation is the creation of code snippets tailored to user needs or to augment code completion processes \citep{feng2020codebert, Li2022}, which includes the development of code LLMs \citep{deepseek2023, zheng2023codegeex}.

\begin{table*}[!hbt]
    \centering
      \caption{Comparison of benchmarks: characterizing existing function-level benchmarks and \textsc{ML-Bench}.}
    \begin{tabular}{cccc}
    \toprule
    \textbf{Bench Name}  & \textbf{Type} & \textbf{Language} & \textbf{\# Samples} \\
    \midrule
    ML-Bench &  Task Execution &  Python \& Bash & 9,641  \\
    HumanEval ~\citep{chen2021codex} &  Function Completion &  Python & 164 \\
    MBPP ~\citep{Austin2021ProgramSW} &  Function Completion &  Python & 1,000  \\
    DS-1000 ~\citep{lai2023ds} &  Function Completion &  Python & 1,000  \\
    APPS ~\citep{hendrycks2021measuring} & Function Completion & Python & 10,000 \\
    \bottomrule
    \end{tabular}
  \label{tab:code_bench}
\end{table*}


\subsection{Agent}

The integration of AI agents in software development is rapidly advancing, with projects like OpenDevin\footnote{https://github.com/OpenDevin/OpenDevin.}, SWE-agent~\citep{yang2024swe}, AutoGen~\citep{wu2023autogen}, and Aider~\citep{aider} showcasing diverse methodologies for augmenting developers' capabilities. OpenDevin$^\S$ leverages open-source technologies to automate coding and debugging, thereby streamlining development workflows. SWE-agent's ACI allows language models to independently tackle software engineering tasks, exhibiting impressive outcomes in benchmark tests. AutoGen's collaborative agent framework melds conversational AI with human and digital tools to automate a breadth of tasks, from programming to problem-solving. Finally, Aider brings LLMs directly into the coding process, enabling true co-editing experiences between AI models like GPT-4o, Claude 3 Opus, and developers within git repositories, enhancing code editing and project management.




\newpage

\section{Dataset Details}
\subsection{Details of Selected GitHub Repositries}
\label{appendix:details_of_selected_github_repositries}


As depicted in Table \ref{table:github_information}, our selection encompasses a range of GitHub repositories varying from language and graph models to multimodal and time-series models. Each repository is chosen for its high-quality contributions to the field and its popularity among the development community, indicated by the number of stars. The repositories, diverse in their updates and number of README files, provide a snapshot of the current landscape of models available on GitHub.

\begin{table*}[h]
  \centering
    \caption{Comprehensive information on selected GitHub repositories. The column labeled "\#README" refers to the number of README files contained within each listed GitHub repository.}
  \small
   \resizebox{\textwidth}{!}{%
  \begin{tabular}{llclcc}
    \toprule

    \textbf{Domain} & \textbf{GitHub} & \textbf{Stars} & \textbf{URL} & \textbf{\#README} & \textbf{Last Updated} \\

    \midrule
    
    \multirow{2}{*}{Language Model} & BERT & 35693 & \url{https://github.com/google-research/bert} & 1  & 2020.03.11 \\
    & Tensor2Tensor & 14280 & \url{https://github.com/tensorflow/tensor2tensor} & 9  & 2023.04.01 \\
    \midrule

    \multirow{1}{*}{Graph Model} & DGL & 12429 & \url{https://github.com/dmlc/dgl} & 154  & 2023.11.16 \\

    \midrule

    \multirow{2}{*}{Biomedical Model} & ESM & 2462 & \url{https://github.com/facebookresearch/esm} & 8  & 2023.06.27 \\
    & MedicalZooPytorch & 1516 & \url{https://github.com/black0017/MedicalZooPytorch} & 21  & 2022.02.07 \\

    \midrule

    \multirow{4}{*}{Vision Model} & PyTorch-GAN & 14947 & \url{https://github.com/eriklindernoren/PyTorch-GAN} & 1  & 2021.01.07 \\
    & Learning3d & 579 & \url{https://github.com/vinits5/learning3d} & 1  & 2023.10.24 \\
    & External-Attention-pytorch & 9949 & \url{https://github.com/xmu-xiaoma666/External-Attention-pytorch} & 1  & 2023.10.25 \\
    & Pytorch-image-models & 30400 & \url{https://github.com/huggingface/pytorch-image-models} & 1  & 2023.11.09 \\
    \midrule

    \multirow{1}{*}{Audio Model} & Muzic & 3866 & \url{https://github.com/microsoft/muzic} & 8  & 2023.12.06 \\

    \midrule

    \multirow{5}{*}{Multi-Modality} & LAVIS & 7300 & \url{https://github.com/salesforce/lavis} & 8  & 2023.09.25 \\
    & IF & 7237 & \url{https://github.com/deep-floyd/if} & 1  & 2023.06.03 \\
    & OPEN-CLIP & 6856 & \url{https://github.com/mlfoundations/open_clip} & 1  & 2023.11.01 \\
    & Stable Diffusion & 31506 & \url{https://github.com/Stability-AI/stablediffusion} & 1  & 2023.03.25 \\
    & Segment-Anything  & 11976 & \url{https://github.com/IDEA-Research/Grounded-Segment-Anything} & 3  & 2023.12.11 \\
    \multirow{1}{*}{Video} & Vid2Vid & 8393 & \url{https://github.com/NVIDIA/vid2vid} & 2  & 2019.07.04 \\
    \midrule

    \multirow{1}{*}{Time-Series Model} & Time-Series-Library & 2670 & \url{https://github.com/thuml/Time-Series-Library} & 1  & 2023.11.10 \\

    \bottomrule
  \end{tabular}
  }
  \label{table:github_information}
  \end{table*}

\subsection{Templates for Diverse Instruction Construction}
\label{appendix:templates_diverse_instruction_construction}

To generate precise instructions, we have developed a range of instruction templates that serve as the foundation for the tasks, ensuring that all necessary arguments are present and accounted for. 
For a detailed look, please refer to the instruction templates outlined in Table \ref{tab:appendix_instruct_template_3}. These templates are instrumental in guiding the generation process, embodying diversity and adaptability while maintaining the integrity of the input information. The templates in the table guide users to regenerate similar instructions with varied expressions while keeping the task requirements intact, promoting diversity in instructional outputs for the model to process and execute.

\begin{table}[h]
\scriptsize
\centering
\caption{Examples of constructed instructions for diverse task templates. "ESMFold" in the context of the table refers to a task related to protein structure prediction using a particular machine learning model or function. The esmfold\_v1() and esm1v\_t33\_650M\_UR90S\_1() mentioned in the instructional templates are names of functions or models within a software package designed to predict the three-dimensional structure of proteins based on their amino acid sequence. In the examples provided in the table, the "ESMFold Structure Prediction" task requires inputs such as a sequence of amino acids and specific model parameters, including learning rates and output file paths. Users are asking for assistance in writing code that utilizes the ESMFold model to predict protein structures and save the results to a specified output file in PDB format, which is a common format for representing 3D structures of large biological molecules.}
\begin{tabular}{p{0.98\linewidth}}
\toprule
\textbf{Templates of Instruction Construction} \\  
\midrule
You need to use your creativity and rewrite each query three times. You need to make sure that all the required arguments are provided. \\

Example 1 \\
Input:I have a data sequence MKTVRQERLKSI, and I want to use esm1v\_t33\_650M\_UR90S\_1() to finish ESMFold Structure Prediction task, set learning rate to 0.000001 then output to output.pdb. Can you help me write some code?

Require arguments:\\
sequence: MKTVRQERLKSI\\
model: esm1v\_t33\_650M\_UR90S\_1()\\
task: ESMFold Structure Prediction task\\
learning rate: 0.000001\\
output path:  output.pdb\\
Output:\\
1. Input data: MKTVRQERLKSI 

model: esm1v\_t33\_650M\_UR90S\_1()  

task: ESMFold Structure Prediction 

output path: output.pdb  

learning rate: 0.000001  

Help me write example code.\\

2. I possess a data sequence with the characters MKTVRQERLKSI, and I'm interested in leveraging the esm1v\_t33\_650M\_UR90S\_1() function to accomplish the ESMFold Structure Prediction task. Please set the learning rate to 0.000001. Subsequently, I'd like to generate an output file named output.pdb. Would you be able to assist me by providing the necessary code?\\
3. I possess a fascinating data sequence bearing the distinctive characters MKTVRQERLKSI. I intend to harness the power of the esm1v\_t33\_650M\_UR90S\_1() function, a formidable tool indeed, to conquer the challenging task of ESMFold Structure Prediction. Please set the learning rate to 0.000001. Once victorious, I aim to put the outcome into a file known as output.pdb. Would you be willing to embark on this coding odyssey with me and craft the necessary script?\\

Example2\\
Input: I want to use esmfol\_v1() to finish the ESMFold Structure Prediction task, set the learning rate to 0.04 and batch size to 5, then output to ./output/data.pdb. Can you help me write some code?\\
Require arguments:\\
model: esmfold\_v1()\\
task: ESMFold Structure Prediction task\\
learning rate: 0.04\\
batch size: 5\\
Output:\\
1.  Do you know ESM? I need to do the ESMFold structure prediction task. I want to use model esmfold\_v1() and set the learning rate to 0.04 and batch size to 5.  Can you help me?\\
2. I have a task at hand involving ESMFold Structure Prediction, and I'm interested in harnessing the esmfold\_v1() model for this purpose. I want to set the learning rate to 0.04 and the batch size to 5. Could you guide me through the process of writing the requisite code?\\
3. I'm looking to employ the esmfold\_v1() function to complete the ESMFold Structure Prediction task, and I would like to set the learning rate to 0.04 and batch size to 5. Could you assist me in crafting the necessary code?\\

You must follow the examples, keep the information unchanged, and write some new instructions. Your output should be diverse. You are encouraged to write long instructions. The instructions should be very different.\\

Input: I'm interested in utilizing ./model/cased\_L-12\_H-768\_A-12 for fine-tuning the SQUAD2.0 task. I plan to establish a train batch size of 128. Could you assist me in writing the necessary code?\\
Require arguments:\\
model path: ./model/cased\_L-12\_H-768\_A-12\\
task: SQUAD2.0\\
batch size: 128\\

Output:\\
1.\\
2.\\
3.\\

\bottomrule
\end{tabular}
\label{tab:appendix_instruct_template_3}
\end{table}

\clearpage

\subsection{Token Length Analysis}

For a detailed examination of the token lengths encountered during our model evaluations, we present two tables (Table~\ref{tab:tokens} and \ref{tab:token_lengths}) comparing the token counts across a variety of experimental settings. The measurements were made using the CodeLlama tokenizer and GPT-3.5 tokenizer.

\begin{table*}[!hbtp]
\centering
\caption{Average input token length of instructions for CodeLlama and GPT-3.5$^*$ in various experimental settings. $^*$We use tiktoken (\href{https://github.com/openai/tiktoken}{https://github.com/openai/tiktoken}) to calculate the token number for GPT-3.5.}
\resizebox{\linewidth}{!}{
\begin{tabular}{llrrrrrrrrrrrr}
\toprule
\multicolumn{2}{c}{\multirow{3}{*}{\textbf{Repository}}} & \multicolumn{6}{c}{\textbf{CodeLlama}}&\multicolumn{6}{c}{\textbf{GPT 3.5}}\\
\cmidrule(lr){3-8} \cmidrule(lr){9-14}
\multicolumn{2}{c}{} & \multicolumn{2}{c}{\textbf{Code}} & \multicolumn{2}{c}{\textbf{Retrieval}} & \multicolumn{2}{c}{\textbf{Oracle}} & \multicolumn{2}{c}{\textbf{Code}} & \multicolumn{2}{c}{\textbf{Retrieval}} & \multicolumn{2}{c}{\textbf{Oracle}} \\
\cmidrule(lr){3-4} \cmidrule(lr){5-6} \cmidrule(lr){7-8} \cmidrule(lr){9-10} \cmidrule(lr){11-12} \cmidrule(lr){13-14}
\multicolumn{2}{c}{} & \textbf{Train} & \textbf{Test} & \textbf{Train} & \textbf{Test} & \textbf{Train} & \textbf{Test} & \textbf{Train} & \textbf{Test} & \textbf{Train} & \textbf{Test} & \textbf{Train} & \textbf{Test}\\
\midrule
\rowcolor[RGB]{238, 255, 255}\multicolumn{14}{c}{\textbf{In-Distribution (ID)}} \\
\multicolumn{2}{l}{\href{https://github.com/dmlc/dgl}{DGL}} & 5,466,687 & 5,466,687 & 312 & 2,603 & 179 & 138 & 4,455,349 & 4,455,349 & 275 & 2,011 & 143 & 110 \\
\multicolumn{2}{l}{\href{https://github.com/google-research/bert}{BERT}} & 138,445 & 138,445 & 401 & 344 & 372 & 375 & 112,104 & 112,104 & 335 & 280 & 287 & 290 \\
\multicolumn{2}{l}{\href{https://github.com/facebookresearch/esm}{ESM}} & 27,107,031 & 27,107,031 & 585 & 438 & 177 & 173 & 22,227,765 & 22,227,765 & 486 & 273 & 139 & 136 \\
\multicolumn{2}{l}{\href{https://github.com/eriklindernoren/PyTorch-GAN}{Py-GAN}} & 146,570 & 146,570 & 532 & 897 & 314 & 314 & 119,454 & 119,454 & 433 & 744 & 268 & 268 \\
\multicolumn{2}{l}{\href{https://github.com/salesforce/lavis}{Lavis}} & 16,827,026 & 16,827,026 & 471 & 401 & 1984 & 1984 & 13,714,026 & 13,714,026 & 372 & 325 & 1547 & 1547 \\
\multicolumn{2}{l}{\href{https://github.com/xmu-xiaoma666/External-Attention-pytorch}{External-Attention (EAP)}} & 449,381 & 449,381 & 1155 & 526 & 105 & 118 & 346,898 & 346,898 & 857 & 412 & 69 & 80 \\
\multicolumn{2}{l}{\href{https://github.com/deep-floyd/if}{If}} & 68,316 & 68,316 & 1390 & 1,642 & 3023 & 3023 & 55,677 & 55,677 & 1119 & 1,330 & 2367 & 2367 \\
\multicolumn{2}{l}{\href{https://github.com/NVIDIA/vid2vid}{vid2vid}} & 146,696 & 146,696 & 408 & 1615 & 556 & 565 & 111,783 & 111,783 & 338 & 481 & 416 & 416 \\
\multicolumn{2}{l}{\href{https://github.com/mlfoundations/open_clip}{OpenCLIP}} & 6,143,829 & 6,143,829 & 415 & 491 & 5420 & 5420 & 5,037,939 & 5,037,939 & 350 & 405 & 4397 & 4397 \\
\multicolumn{2}{l}{\href{https://github.com/thuml/Time-Series-Library}{TSL}} & 337,114 & 337,114 & 382 & 902 & 345 & 345 & 273,062 & 273,062 & 315 & 731 & 276 & 276 \\
\midrule
\rowcolor[RGB]{238, 255, 255}\multicolumn{14}{c}{\textbf{Out-Of-Distribution (OOD)}} \\
\multicolumn{2}{l}{\href{https://github.com/IDEA-Research/Grounded-Segment-Anything}{Grounded-SAM}} & \slash & 16,726,416 & \slash & 898 & \slash & 164 & \slash & 13,715,662 & \slash & 754 & \slash & 113 \\
\multicolumn{2}{l}{\href{https://github.com/huggingface/pytorch-image-models}{Py-IM}} & \slash & 5,608,249 & \slash & 8,025 & \slash & 89 & \slash & 4,542,681 & \slash & 6,415 & \slash & 68 \\
\multicolumn{2}{l}{\href{https://github.com/microsoft/muzic}{muzic}} & \slash & 13,325,828 & \slash & 616 & \slash & 83 & \slash & 10,860,549 & \slash & 507 & \slash & 64 \\
\multicolumn{2}{l}{\href{https://github.com/vinits5/learning3d}{Learning3D}} & \slash & 320,157 & \slash & 640 & \slash & 50 & \slash & 256,110 & \slash & 596 & \slash & 45 \\
\multicolumn{2}{l}{\href{https://github.com/Stability-AI/stablediffusion}{SD}} & 258,096 & \slash & 501 & \slash & 234 & \slash & 209,058 & \slash & 412 & \slash & 183 & \slash \\
\multicolumn{2}{l}{\href{https://github.com/black0017/MedicalZooPytorch}{MedZooPy}} & 2,701,443 & \slash & 1,302 & \slash & 133 & \slash & 2,150,168 & \slash & 1,101 & \slash & 99 & \slash \\
\multicolumn{2}{l}{\href{https://github.com/brightmart/text_classification}{TCL}} & 18,696,614 & \slash & 345 & \slash & 116 & \slash & 15,114,250 & \slash & 291 & \slash & 96 & \slash \\
\multicolumn{2}{l}{\href{https://github.com/tensorflow/tensor2tensor}{Tensor2Tensor}} & 4,598,727 & \slash & 501 & \slash & 192 & \slash & 3,678,980 & \slash & 432 & \slash & 153 & \slash \\
\bottomrule
\end{tabular}
}
\label{tab:tokens}
\end{table*}

\begin{table*}[!hbt]
\centering
\caption{Average output token length of code for GPT-3.5$^*$ and CodeLlama to generate across different datasets (Train Set, Test Set, 1/4 Test Set) for various repositories, separated by Python Code and Bash Script. $^*$We use tiktoken (\href{https://github.com/openai/tiktoken}{https://github.com/openai/tiktoken}) to calculate the token number for GPT-3.5.}
\small
\resizebox{\linewidth}{!}{
\begin{tabular}{lrrrrrrrrrrrr}
\toprule
\multirow{3}{*}{\textbf{Repository}} & \multicolumn{4}{c}{\textbf{Train Set}} & \multicolumn{4}{c}{\textbf{Test Set}} & \multicolumn{4}{c}{\textbf{1/4 Test Set}} \\
\cmidrule(lr){2-5} \cmidrule(lr){6-9} \cmidrule(lr){10-13}
& \multicolumn{2}{c}{\textbf{GPT-3.5}} & \multicolumn{2}{c}{\textbf{CodeLlama}} & \multicolumn{2}{c}{\textbf{GPT-3.5}} & \multicolumn{2}{c}{\textbf{CodeLlama}} & \multicolumn{2}{c}{\textbf{GPT-3.5}} & \multicolumn{2}{c}{\textbf{CodeLlama}} \\
\cmidrule(lr){2-3} \cmidrule(lr){4-5} \cmidrule(lr){6-7} \cmidrule(lr){8-9} \cmidrule(lr){10-11} \cmidrule(lr){12-13}
& \textbf{Python} & \textbf{Bash} & \textbf{Python} & \textbf{Bash} & \textbf{Python} & \textbf{Bash} & \textbf{Python} & \textbf{Bash} & \textbf{Python} & \textbf{Bash} & \textbf{Python} & \textbf{Bash} \\
\midrule
\rowcolor[RGB]{238, 255, 255}\multicolumn{13}{c}{\textbf{In-Distribution (ID)}} \\
\href{https://github.com/dmlc/dgl}{DGL} & \slash & 21.15 & \slash & 28.05 & \slash & 18.24 & \slash & 24.33 & \slash & 21.60 & \slash & 28.40 \\
\href{https://github.com/google-research/bert}{BERT} & \slash & 121.98 & \slash & 181.60 & \slash & 120.14 & \slash & 179.36 & \slash & 127.67 & \slash & 189.50 \\
\href{https://github.com/facebookresearch/esm}{ESM} & 142.79 & 37.80 & 183.84 & 52.44 & 127.50 & 37.47 & 167.50 & 52.40 & 127.00 & 40.00 & 167.00 & 54.25 \\
\href{https://github.com/eriklindernoren/PyTorch-GAN}{Py-GAN} & \slash & 28.63 & \slash & 43.25 & \slash & 27.30 & \slash & 41.10 & \slash & 27.00 & \slash & 40.88 \\
\href{https://github.com/salesforce/lavis}{Lavis} & 222.95 & 36.05 & 313.97 & 51.72 & 211.30 & 34.75 & 300.57 & 49.25 & 187.33 & 37.00 & 267.00 & 51.00 \\
\href{https://github.com/xmu-xiaoma666/External-Attention-pytorch}{EAP} & 170.87 & \slash & 239.68 & \slash & 121.63 & \slash & 174.96 & \slash & 146.20 & \slash & 205.60 & \slash \\
\href{https://github.com/deep-floyd/if}{If} & 243.47 & 160.00 & 325.42 & 201.00 & 272.19 & \slash & 362.57 & \slash & 269.33 & \slash & 361.83 & \slash \\
\href{https://github.com/NVIDIA/vid2vid}{vid2vid} & \slash & 85.65 & \slash & 112.67 & \slash & 79.85 & \slash & 104.85 & \slash & 63.25 & \slash & 84.75 \\
\href{https://github.com/mlfoundations/open_clip}{OpenCLIP} & 859.31 & \slash & 1236.63 & \slash & 839.55 & \slash & 1207.91 & \slash & 913.33 & \slash & 1313.33 & \slash \\
\href{https://github.com/thuml/Time-Series-Library}{TSL} & \slash & 152.98 & \slash & 205.82 & \slash & 151.07 & \slash & 204.71 & \slash & 152.75 & \slash & 207.00 \\
\midrule
\rowcolor[RGB]{238, 255, 255}\multicolumn{13}{c}{\textbf{Out-Of-Distribution (OOD)}} \\
\href{https://github.com/huggingface/pytorch-image-models}{Py-IM} & \slash & \slash & \slash & \slash & \slash & 37.40 & \slash & 53.00 & \slash & 26.00 & \slash & 34.00 \\
\href{https://github.com/vinits5/learning3d}{Learning3D} & \slash & \slash & \slash & \slash & \slash & 28.59 & \slash & 41.00 & \slash & 27.75 & \slash & 41.00 \\
\href{https://github.com/microsoft/muzic}{muzic} & \slash & \slash & \slash & \slash & \slash & 26.72 & \slash & 38.72 & \slash & 14.40 & \slash & 21.80 \\
\href{https://github.com/IDEA-Research/Grounded-Segment-Anything}{Grounded-SAM} & \slash & \slash & \slash & \slash & 177.88 & 48.08 & 271.25 & 67.75 & 177.67 & 62.00 & 271.67 & 88.50 \\
\midrule
\textbf{Average (ID)} & 327.88 & 80.53 & 459.91 & 109.57 & 314.43 & 66.97 & 442.70 & 93.71 & 328.64 & 67.04 & 462.95 & 93.68 \\
\textbf{Average (OOD)} & \slash & \slash & \slash & \slash & 177.88 & 35.20 & 271.25 & 50.12 & 177.67 & 32.54 & 271.67 & 46.33 \\
\textbf{Total Average} & 327.88 & 80.53 & 459.91 & 109.57 & 291.79 & 60.12 & 414.09 & 84.15 & 303.64 & 59.04 & 431.07 & 84.11 \\
\bottomrule
\end{tabular}
}
\label{tab:token_lengths}
\end{table*}


\subsection{Detailed Analysis of Task Variety and Instructional Depth}

To provide a clearer understanding of the scope and diversity within \textsc{ML-Bench}, Table \ref{tab:task_num} offers a detailed enumeration of the different types of tasks as well as an analysis of the intricacies involved in the instructions that accompany them. Each task category represents a unique section of our dataset, with Multi-Modality tasks taking the lead with 4,732 instances.
 Time-series and Text-related tasks follow suit with 1,478 and 1,475 instances, respectively, indicating a substantial focus on these areas as well. The numbers are counted by our 8 annotators.

Further linguistic analysis revealed the instruction sets' complexity, with an average token length per instruction measuring 80.4 and a maximum token length reaching up to 216 tokens. 
Additionally, the instruction edit distance—an indicator of linguistic diversity—averages 258.7 tokens within similar tasks and 302.1 tokens across different tasks, underlining the variety and broad coverage of scenarios that \textsc{ML-Bench} encompasses. 

\begin{table*}[!hbpt]
\centering
\caption{Task distribution and instruction complexity.}
\resizebox{0.5\linewidth}{!}{
\begin{tabular}{lr}
\toprule
\textbf{Task} & \textbf{Number} \\ \midrule
 \quad\mbox{-} GNN & 608 \\
 \quad\mbox{-} Text & 1475 \\
 \quad\mbox{-} Molecular & 649 \\
 \quad\mbox{-} Image-GAN & 1189 \\
 \quad\mbox{-} Multi-Modality & 4732 \\
 \quad\mbox{-} Video & 75 \\
 \quad\mbox{-} Time-series & 1478\\
 \quad\mbox{-} Attention Usage & 127 \\
 \quad\mbox{-} Medical & 805\\
 \quad\mbox{-} 3D & 264\\
 \quad\mbox{-} Music & 704\\
\midrule
Average token length per instruction & 80.4 \\
Max token length in instruction & 216 \\
Instruction edit distance among the same task & 258.7\\
Instruction edit distance across tasks & 302.1\\
\bottomrule
\end{tabular}
}
\label{tab:task_num}
\end{table*}




    

\subsection{Error Analysis for Each Repository}

\begin{figure*}[!hbtp]
\centering
\includegraphics[width=0.9\textwidth]{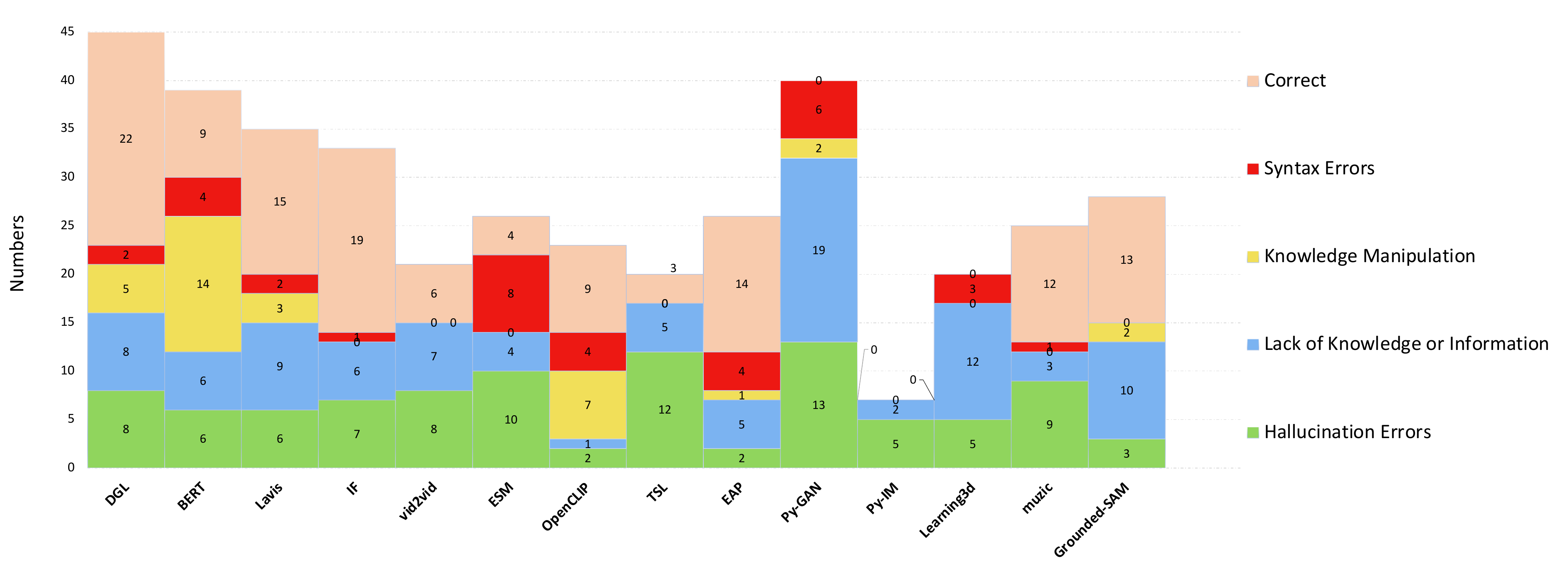}
\caption{Using the Oracle setup, we ran GPT-4 for five iterations and tallied the number of errors across different repositories to provide an error analysis specific to each repository.}
\label{fig:repo_analysis}
\end{figure*}

Figure \ref{fig:repo_analysis} illustrates the distribution of errors made by GPT-4 across 14 repositories, categorized as per the error types described in the main text. The analysis was conducted within the context of the \textsc{ML-LLM-Bench}, specifically under the Oracle setting.

\subsection{Error Analysis for \textsc{ML-Agent-Bench}}

\begin{figure*}[!hbtp]
\centering
\includegraphics[width=0.9\textwidth]{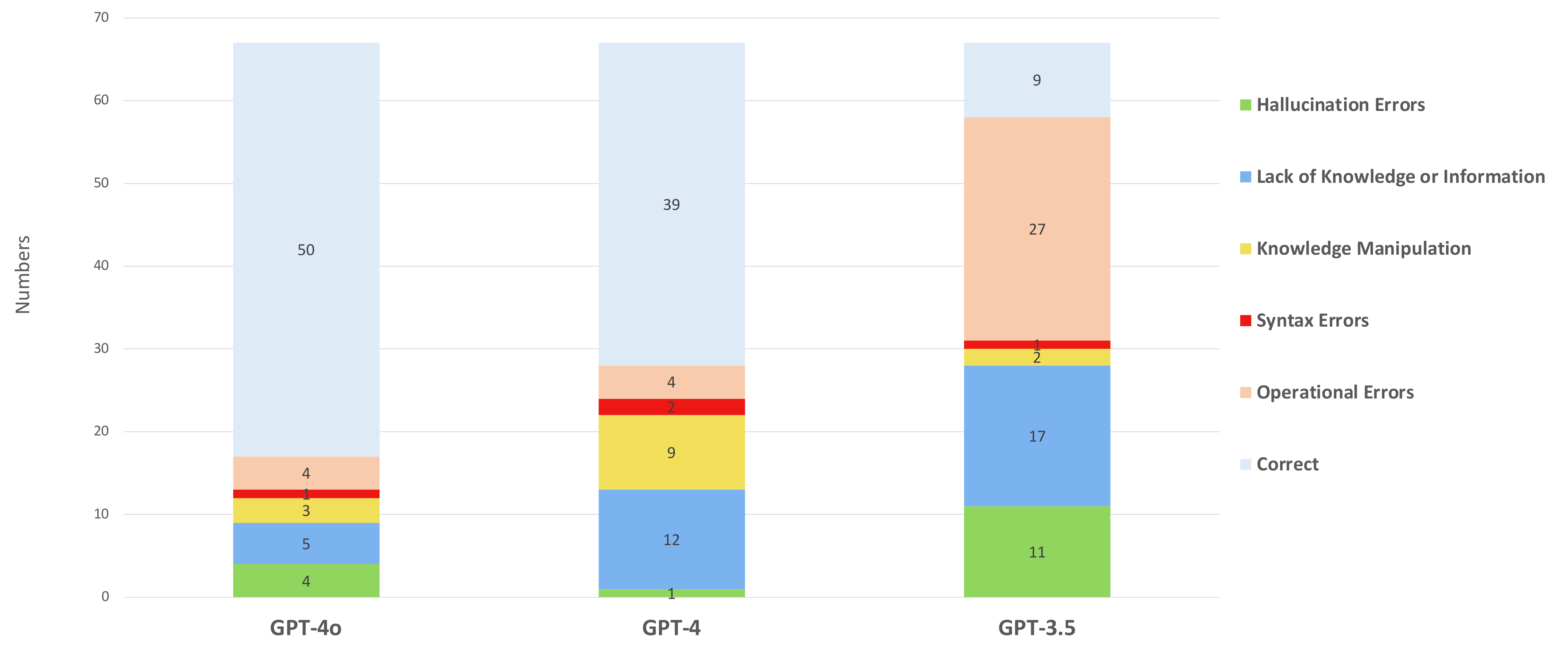}
\caption{The error analysis for the OpenDevin framework, utilizing various base models on \textsc{ML-Agent-Bench}. Notably, Operational Error is a category unique to \textsc{ML-Agent-Bench} due to the agents' need to interact dynamically with the coding environment.}
\label{fig:ml_agent_error_analysis}
\end{figure*}

Figure \ref{fig:ml_agent_error_analysis} illustrates the distribution of errors made by OpenDevin, categorized as per the error types described in the main text. The analysis was conducted within the context of the \textsc{ML-Agent-Bench}.

\clearpage

\subsection{An Example of \textsc{ML-LLM-Bench}}
\label{app:input_output}

\begin{figure*}[h]
\centering
\includegraphics[width=\textwidth]{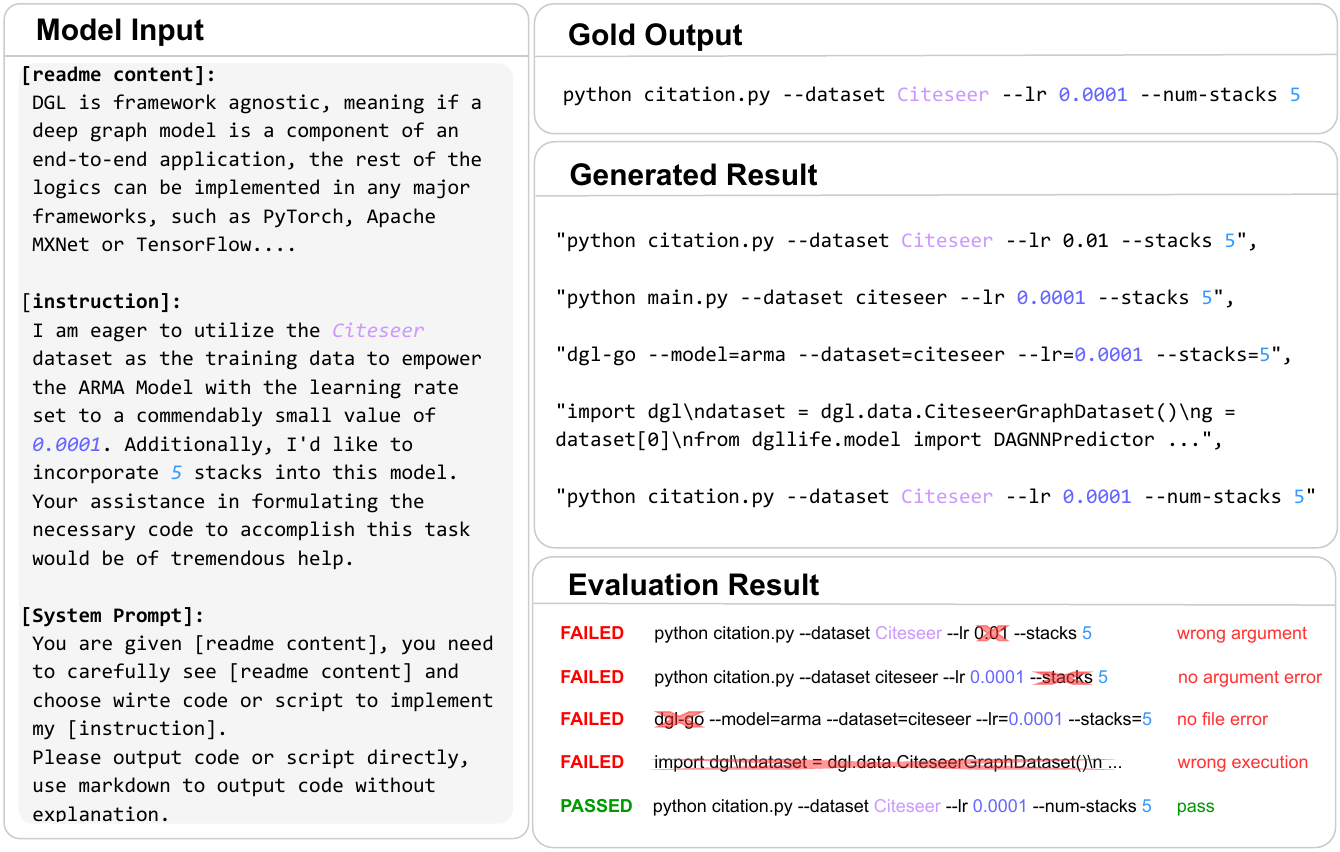}
\caption{The illustrative set of \textbf{input-output pairs, gold output, and execution results}, accompanied by precision metrics: \textbf{Pass@1=0 and Pass@5=1}. Various colors within the instructions signify different parameters.}
\label{fig:input_output_eval}
\end{figure*}

Figure \ref{fig:input_output_eval} shows a case from \textsc{ML-LLM-Bench} includes a diagram that provides a view of the model's inputs, the reference, multiple outputs generated by the model, and an error analysis with results for each output.

\clearpage

\subsection{An Example of \textsc{ML-Agent-Bench}}
In this section, we present detailed examples of each error type in \textsc{ML-Agent-Bench}:
\begin{enumerate}
    \item Hallucination Errors (E1): Table \ref{table:task_error_E1}
    \item Lack of Knowledge or Information (E2): Table \ref{table:task_error_E2}
    \item Knowledge Manipulation (E3): Table \ref{table:task_error_E3}
    \item Syntax Errors (E4): Table \ref{table:task_error_E4}
    \item Operational Error (E5): Table \ref{table:task_error_E5}
\end{enumerate}

\input{agent_tables/example_E1}
\input{agent_tables/example_E2}
\input{agent_tables/example_E3}
\input{agent_tables/example_E4}
\input{agent_tables/example_E5}

\clearpage

\newpage

\section{Experiment Details}

\subsection{Efficacy of BM25 Retrieval on Repository-Level Code}
\label{app:bleu_score_cutoff_date}

When evaluating the relevance of retrieved code segments within repository-level tasks, we utilized the BLEU-1 score as a metric to measure the similarity between hand-annotated "Oracle" segments and those extracted by the model using BM25 retrieval. "Oracle" segments are those identified by annotators as most pertinent, serving as the foundational evidence required for the task at hand. As depicted in Table \ref{tab:bm25_hit_proba}, the resulting BLEU scores indicate a low degree of correlation, suggesting that the retrieval segments identified by BM25 are significantly dissimilar to the Oracles crafted by human annotators. This finding is demonstrative of BM25's limitations in effectively identifying the most relevant content for repository-scale code, as evidenced by the low BLEU scores.

\begin{table*}[h]
\centering
\caption{The BLEU-1 scores between Oracles and the Retrieval content.}
\resizebox{0.5\linewidth}{!}{%
\begin{tabular}{c c c c}
\toprule
& \textbf{ID-train} & \textbf{OOD-train} & \textbf{\benchname} \\
\midrule
BLEU score & 0.0112 & 0.0087 & 0.0082 \\
\bottomrule
\end{tabular}
}
\label{tab:bm25_hit_proba}
\hfill
\end{table*}



 



\subsection{Information Lossing due to Truncation}
\label{appendix:information_loss}

It is reasonable that truncation may lead to information missing, but it is worth noting that only in the Code setting for the open-source models does the input of README files need to be truncated to 8k, which is inevitable because of the input length limitation. However, only a small number of README files need to be truncated. To qualitatively present the information loss percentage due to truncation, we present the percentage of losing critical information during truncation in Table~\ref{tab:losing-information}. Note that all the results are manually examined. We can identify that only 5 repositories lose critical information after truncating the README files.

\begin{table}[!h]
\centering
\caption{The percentage of losing critical information due to truncation.}
\small
\begin{tabular}{llc}
\toprule
\multicolumn{2}{c}{\textbf{Repos}} & \textbf{Proportion of losing information (\%)} \\
\midrule
\multicolumn{2}{l}{vid2vid} & 0 \\
\multicolumn{2}{l}{If} & 0 \\
\multicolumn{2}{l}{DGL} & 0 \\
\multicolumn{2}{l}{Py-GAN} & 33.3 \\
\multicolumn{2}{l}{ESM} & 11.76 \\
\multicolumn{2}{l}{BERT} & 100 \\
\multicolumn{2}{l}{OpenCLIP} & 0 \\
\multicolumn{2}{l}{Lavis} & 0 \\
\multicolumn{2}{l}{TSL} & 0 \\
\multicolumn{2}{l}{EAP} & 75 \\
\multicolumn{2}{l}{Grounded-SAM} & 0 \\
\multicolumn{2}{l}{Py-IM} & 20 \\
\multicolumn{2}{l}{muzic} & 0 \\
\multicolumn{2}{l}{Learning3d} & 0 \\
\multicolumn{2}{l}{SD} & 0 \\
\multicolumn{2}{l}{MedZooPy} & 0 \\
\multicolumn{2}{l}{TCL} & 0 \\
\multicolumn{2}{l}{Tensor2Tensor} & 0 \\
\multicolumn{2}{l}{Total} & 0 \\
\bottomrule
\end{tabular}
\label{tab:losing-information}
\end{table}





\clearpage

\subsection{Examples of Input-output of each GitHub Repository}
In this section, we present detailed examples of the input and output of each GitHub Repo in Tab.\ref{table:example_as} to Tab.\ref{table:example_vid2vid}. The corresponding repository for each table is shown below:

\begin{enumerate}
    \item External-Attention: Table ~\ref{table:example_as}
    \item BERT: Table ~\ref{table:example_bert}
    \item Deep learning on graphs: Table ~\ref{table:example_DGL}
    \item Evolutionary scale modeling: Table ~\ref{table:example_esm}
    \item Grounded-Segment-Anything: Table ~\ref{table:example_gsa}
    \item DeepFloyd IF: Table ~\ref{table:example_if}
    \item Language-Vision Intelligence: Table ~\ref{table:example_lavis}
    \item Deep learning on 3D point clouds data: Table ~\ref{table:example_learning3D}
    \item 3D multi-modal medical image segmentation library: Table ~\ref{table:example_medicalzoo}
    \item Music understanding and generation: Table ~\ref{table:example_muzic}
    \item Implementation of OpenAI's CLIP: Table ~\ref{table:example_openCLIP}
    \item Generative Adversarial Network varieties: Table ~\ref{table:example_pyGAN}
    \item PyTorch Image Models: Table ~\ref{table:example_PyIM}
    \item Stable diffusion: Table ~\ref{table:example_SD}
    \item Text classification: Table ~\ref{table:example_TC}
    \item Tensor2Tensor: Table ~\ref{table:example_tensor2}
    \item deep time series analysis: Table ~\ref{table:example_tsl}
    \item Video-to-video translation: Table ~\ref{table:example_vid2vid}
\end{enumerate}

\subfile{github_tables/example_AS}
\subfile{github_tables/example_bert}
\subfile{github_tables/example_DGL}
\subfile{github_tables/example_ESM}
\subfile{github_tables/example_gsa}
\subfile{github_tables/example_IF}
\subfile{github_tables/example_lavis}
\subfile{github_tables/example_learning3D}

\subfile{github_tables/example_MedicalZoo}
\subfile{github_tables/example_muzic}
\subfile{github_tables/example_openCLIP}
\subfile{github_tables/example_PyGAN}
\subfile{github_tables/example_PyIM}
\subfile{github_tables/example_SD}
\subfile{github_tables/example_TC}
\subfile{github_tables/example_tensor2}
\subfile{github_tables/example_tsl}
\subfile{github_tables/example_vid2vid}

\clearpage

\subsection{BM25}

\textbf{Retrieval}: 
In this setting, we apply BM25 retriever to get the instruction-related segment in README files.
We set the BM25 retriever's retrieval span at 10 sentences using the NLTK sentence tokenizer because the average length of an Oracle is 9.5. 

\subsection{Results Across Different GitHub Repositories}
\begin{table*}[h]
\centering
\caption{This table shows the Pass@5 scores of GPT-4, and Claude on the 1⁄4 set (*) and GPT 3.5 and Claude on the full set across different GitHub repositories.
\textbf{Oracle}, \textbf{Code} and \textbf{BM25} represents 
Oracle, Code, and Retrieval settings.}
\resizebox{\linewidth}{!}{
\begin{tabular}{lcccccccccccccc}
\toprule
\multicolumn{2}{c}{\multirow{2}{*}{\textbf{Repository}}}  & \multicolumn{3}{c}{\textbf{GPT-4*}} & \multicolumn{3}{c}{\textbf{Claude *}} & \multicolumn{3}{c}{\textbf{GPT 3.5}} & \multicolumn{3}{c}{\textbf{Claude }}\\
\cmidrule(lr){3-5} \cmidrule(lr){6-8} \cmidrule(lr){9-11} \cmidrule(lr){12-14} 
\multicolumn{2}{c}{}  & \textbf{Oracle}& \textbf{Code}  & \textbf{BM25} & \textbf{Oracle} & \textbf{Code}  & \textbf{BM25} & \textbf{Oracle} & \textbf{Code}  & \textbf{BM25} & \textbf{Oracle} & \textbf{Code}  & \textbf{BM25}  \\
\midrule
\multicolumn{2}{l}{DGL}   & \textbf{80.00}  & 60.00   & 60.00  & 40.00 & 20.00  & 80.00 & \textbf{47.62} & 23.81  & 23.81 & 28.57 & 19.05 & 14.29 \\
\multicolumn{2}{l}{BERT}  & 50.00   & 50.00  & 16.67 & 0.00 & \textbf{80.00}   & 16.67 & \textbf{22.73}& 13.63 & 13.63 & 0.00 & 4.54 & 0.00\\
\multicolumn{2}{l}{Lavis}   & 42.86   & 71.43  & 42.86 & 57.14 & \textbf{85.71}  & 14.29 & 55.56 & \textbf{70.37}  & 51.85 & 51.85 & 59.26 & 29.63 \\
\multicolumn{2}{l}{If} & \textbf{100.00}     & \textbf{100.00}    & 33.33 & \textbf{100.00} & 0.00  & 13.33 & 71.43 & 61.90  & 52.38 & 71.43 & \textbf{76.19} & 52.38 \\
\multicolumn{2}{l}{vid2vid}  & 50.00  & \textbf{75.00}  & 50.00 & 0.00 & 25.00  & 50.00 & \textbf{92.31} & 76.92 & 69.23 & 76.92 & 38.46 & 15.38 \\
\multicolumn{2}{l}{ESM}  & 60.00 & 0.00  & 80.00 & 0.00 & \textbf{100.00}  & 20.00 & 47.06 & 29.41 & \textbf{58.82} & 5.88 & 11.76 & 11.76\\
\multicolumn{2}{l}{OpenCLIP} & \textbf{66.67} & \textbf{66.67}  & \textbf{66.67} & \textbf{66.67} & \textbf{66.67}  & 0.00 & \textbf{63.63} & 36.36 &  54.55 & \textbf{63.63} & \textbf{63.63} & 45.46 \\
\multicolumn{2}{l}{TSL}  & 25.00  & 25.00  & 0.00  & 25.00 & 0.00  & 0.00 & \textbf{14.29} & \textbf{14.29} & 0.00 & 7.14 & 7.14 & 0.00 \\
\multicolumn{2}{l}{EAP}  & \textbf{100.00}  & 80.00 & 0.00  & 100.00 & 20.00 & 80.00 & 66.66 & 70.83 & 33.33 & 70.83 & \textbf{83.33} & 20.83 \\
\multicolumn{2}{l}{Py-GAN}   & 0.00  & \textbf{12.50}  & 0.00 & 0.00 & \textbf{12.50}  & 0.00 & \textbf{6.67} & 0.00  & 0.00 & 0.00 & 0.00 & 0.00\\
\multicolumn{2}{l}{Py-IM}  & 0.00 & 0.00 & 0.00 & 0.00 & 0.00 & 0.00 & \textbf{20.00} & 0.00  & 0.00 & 0.00 & 0.00 & 0.00 \\
\multicolumn{2}{l}{Learning3d} & 0.00 & 0.00 & 0.00 & \textbf{25.00} & 0.00 & \textbf{25.00} & 23.53 & \textbf{47.06}& 35.29 & 17.65 & 0.00 & 0.00 \\
\multicolumn{2}{l}{muzic}  & \textbf{80.00} & 60.00 & 40.00 & 60.00 & 20.00 & 20.00 & 66.67 & \textbf{72.22}  & 61.11 & 38.89 & 33.33 & 33.33 \\
\multicolumn{2}{l}{Grounded-SAM} & \textbf{60.00} & \textbf{60.00} & 20.00 & 0.00 & 0.00 & 0.00 & 0.00 & \textbf{20.00} & 0.00 & 5.00 & 35.00 & 10.00 \\
\multicolumn{2}{l}{Total} & \textbf{48.53} & 45.59 & 27.94 & 34.25 & 35.61 & 20.55 & 36.92 & 35.39 & 22.69 & 30.38 & 32.31 & 16.92 \\
\bottomrule
\end{tabular}
}
\label{tab:repo_model_results}
\end{table*}

\section{Contribution}

The team was led by Xiangru Tang, Yuliang Liu, and Zefan Cai, who not only ran the experiments and designed the models but also played a role in the manuscript preparation. Yanjun also contributed a lot to paper writing. Yanjun Shao, Junjie Lu, and Yichi Zhang supported executing the experimental work. The construction of the dataset, which included tasks such as data annotation and cleansing, was undertaken by Zexuan Deng, Helan Hu, Kaikai An, Ruijun Huang, Shuzheng Si, Sheng Chen, Haozhe Zhao, and Liang Chen. Advisors, e.g., Yan Wang, Tianyu Liu, Zhiwei Jiang, Baobao Chang, Yin Fang, Yujia Qin, Wangchunshu Zhou, Yilun Zhao, Arman Cohan, and Mark Gerstein significantly contributed through discussions.

%% file: agent_tables/example_E1.tex
\begin{table*}[!hbtp]
\centering
\begin{tabular}{p{0.98\linewidth}}
\toprule
\hline
\rowcolor{gray!20}
\textbf{Instruction:}\\
\hline
I have a digital image titled \texttt{example.jpg} that I hope to transform into an ink-wash-styled masterpiece with the IF-I-M-v1.0 model. Can you guide me to write the code to transform the image accordingly?\\
\textbf{Steps:}\\
\hline
\rowcolor{green!10}
\textbf{Navigate to the repository}\\
\hline
\begin{lstlisting}[numbers=none, language=bash, backgroundcolor=\color{green!10}]
cd /workspace/if/
\end{lstlisting}\\
\hline
\rowcolor{red!10}
\textbf{Attempt to create the \texttt{run.sh} script in the wrong directory}\\
\hline
The \texttt{run.sh} script is mistakenly attempted to be created in \texttt{/usr/bin} instead of the specified path:
\begin{lstlisting}[numbers=none, language=bash, backgroundcolor=\color{red!10}]
echo "#!/bin/bash" > /usr/bin/run.sh
echo "python if_im.py --model_name IF-I-M-v1.0 --input_path example.jpg --output_path output.jpg" >> /usr/bin/run.sh
chmod +x /usr/bin/run.sh
\end{lstlisting}\\
\hline
\rowcolor{orange!10}
\textbf{Error: Permission denied}\\
\hline
The script creation fails due to insufficient permissions to write in \texttt{/usr/bin}:
\begin{lstlisting}[numbers=none, language=bash, backgroundcolor=\color{orange!10}]
bash: /usr/bin/run.sh: Permission denied
\end{lstlisting}\\
\hline
\rowcolor{red!10}
\textbf{Error: Script not created and task not executed}\\
\hline
The task is not executed because the script could not be created in the specified path:
\begin{lstlisting}[numbers=none, backgroundcolor=\color{red!10}]
The run.sh script could not be created in /usr/bin due to permission issues. The task in /workspace/if/ was not executed.
\end{lstlisting}\\
\bottomrule
\end{tabular}
\caption{An example of E1 error. Task execution attempt for Zero-shot Image-to-Image Translation. The task could not be completed due to an attempt to create the script in \texttt{/usr/bin}, which resulted in permission denial, and the intended task was not executed.}
\label{table:task_error_E1}
\end{table*}

%% file: agent_tables/example_E2.tex
\begin{table*}[!hbtp]
\centering
\begin{tabular}{p{0.98\linewidth}}
\toprule
\hline
\rowcolor{gray!20}
\textbf{Instruction:}\\
\hline
Can you assist me in writing the code to extract 5 designs from the structure \texttt{Fam20A.pdb}, specifically from chain B, at a temperature of 8?\\

\hline
\rowcolor{gray!20}
\textbf{Steps:}\\
\hline
\rowcolor{green!10}
\textbf{Navigate to the repository}\\
\hline
\begin{lstlisting}[numbers=none, language=bash, backgroundcolor=\color{green!10}]
cd /workspace/esm/
\end{lstlisting}\\
\hline
\rowcolor{red!10}
\textbf{Create the \texttt{run.sh} script}\\
\hline
The \texttt{run.sh} script is created with the following content:
\begin{lstlisting}[numbers=none, language=bash, backgroundcolor=\color{red!10}]
#!/bin/bash

python main.py --input Fam20A.pdb --chain B --temperature 8 --output designs.txt --num_designs 5
\end{lstlisting}\\
\hline
\rowcolor{green!10}
\textbf{Run the script}\\
\hline
\begin{lstlisting}[numbers=none, language=bash, backgroundcolor=\color{green!10}]
./run.sh
\end{lstlisting}\\
\hline
\rowcolor{orange!10}
\textbf{Error: Missing \texttt{main.py} script}\\
\hline
The execution of \texttt{run.sh} fails due to the usage of \texttt{main.py} instead of \texttt{extract\_designs.py}:
\begin{lstlisting}[numbers=none, language=bash, backgroundcolor=\color{orange!10}]
python: cannot open file main.py: [Errno 2] No such file or directory
\end{lstlisting}\\
\hline
\rowcolor{red!10}
\textbf{Check for the missing script}\\
\hline
The user is asked to ensure that the \texttt{main.py} script is present in the repository:
\begin{lstlisting}[numbers=none, backgroundcolor=\color{red!10}]
Please make sure that the main.py script is available in the /workspace/esm/ repository.
\end{lstlisting}\\
\hline
\rowcolor{orange!10}
\textbf{Task cannot be completed}\\
\hline
The task cannot be completed without the \texttt{main.py} script:
\begin{lstlisting}[numbers=none, backgroundcolor=\color{orange!10}]
I cannot complete the task as requested without the missing script.
\end{lstlisting}\\
\bottomrule
\end{tabular}
\caption{An example of E2 error. Task execution attempt for extracting designs from \texttt{Fam20A.pdb}. The task could not be completed due to the misinformation of \texttt{main.py} script. The LLM agent should use \texttt{extract\_designs.py} instead.}
\label{table:task_error_E2}
\end{table*}

%% file: agent_tables/example_E3.tex
\begin{table*}[!hbtp]
\centering
\begin{tabular}{p{0.98\linewidth}}
\toprule
\hline
\rowcolor{gray!20}
\textbf{Instruction:}\\
\hline
Looking to use the Cluster GAN model for a project. Hoping to set up a learning rate of 0.0001 and plan on running it for 100 epochs. I would like the batch size to be 16 and to keep the image size at 128. Can you help me craft some code?\\
\hline
\rowcolor{gray!20}
\textbf{Steps:}\\
\hline
\rowcolor{red!10}
\textbf{Create the \texttt{run.sh} script}\\
\hline
The \texttt{run.sh} script is created with the following content:
\begin{lstlisting}[numbers=none, language=bash, backgroundcolor=\color{red!10}]
#!/bin/bash

python main.py --model cluster_gan --learning_rate 0.0001 --num_epochs 100 --batch_size 16 --image_size 128
\end{lstlisting}\\
\hline
\rowcolor{green!10}
\textbf{Run the script}\\
\hline
\begin{lstlisting}[numbers=none, language=bash, backgroundcolor=\color{green!10}]
cd /workspace/PyTorch-GAN/implementations/cluster_gan/
chmod +x run.sh
./run.sh
\end{lstlisting}\\
\hline
\rowcolor{orange!10}
\textbf{Error: Missing \texttt{main.py} script}\\
\hline
The execution of \texttt{run.sh} fails due to the missing \texttt{main.py} script:
\begin{lstlisting}[numbers=none, language=bash, backgroundcolor=\color{orange!10}]
python: cannot open file /workspace/PyTorch-GAN/implementations/cluster_gan/main.py: [Errno 2] No such file or directory
\end{lstlisting}\\
\hline
\rowcolor{red!10}
\textbf{Update \texttt{run.sh} script to use \texttt{clustergan.py}}\\
\hline
The \texttt{run.sh} script is updated with the following content:
\begin{lstlisting}[numbers=none, language=bash, backgroundcolor=\color{red!10}]
#!/bin/bash

python clustergan.py --model cluster_gan --learning_rate 0.0001 --num_epochs 100 --batch_size 16 --image_size 128
\end{lstlisting}\\
\hline
\rowcolor{green!10}
\textbf{Run the updated script}\\
\hline
\begin{lstlisting}[numbers=none, language=bash, backgroundcolor=\color{green!10}]
cd /workspace/PyTorch-GAN/implementations/cluster_gan/
chmod +x run.sh
./run.sh
\end{lstlisting}\\
\hline
\rowcolor{orange!10}
\textbf{Error: Unrecognized arguments in \texttt{clustergan.py}}\\
\hline
The execution of \texttt{run.sh} fails due to unrecognized arguments in \texttt{clustergan.py}:
\begin{lstlisting}[numbers=none, language=bash, backgroundcolor=\color{orange!10}]
usage: clustergan.py [-h] [-n N_EPOCHS] [-b BATCH_SIZE] [-i IMG_SIZE]
                     [-d LATENT_DIM] [-l LEARNING_RATE] [-c N_CRITIC] [-w]
clustergan.py: error: unrecognized arguments: --model cluster_gan --learning_rate 0.0001 --num_epochs 100 --image_size 128
\end{lstlisting}\\
\bottomrule
\end{tabular}
\caption{An example of E3 error. Task execution attempt for Cluster GAN model. The task could not be completed due to unrecognized arguments in \texttt{clustergan.py}. The LLM agent should have ensured the script accepts the required arguments.}
\label{table:task_error_E3}
\end{table*}

%% file: agent_tables/example_E4.tex
\begin{table*}[!hbtp]
\centering
\begin{tabular}{p{0.98\linewidth}}
\toprule
\hline
\rowcolor{gray!20}
\textbf{Instruction:}\\
\hline
Please complete the Machine Learning task in the following repository: esm\\
The task is: Sample sequences design for certain structure\\
Can you assist me in writing the code to extract 3 structure examples from the inverse\_folding data file \texttt{5YH2.pdb}? I specifically need designs with chain A at a temperature of 5. The output should be saved in the \texttt{model\_structure.fasta} file.\\
You should create a script named \texttt{run.sh} under the specified path in the repo to run the task.\\
You can find the task repo at: \texttt{/workspace/esm/}\\
You should terminate the subprocess after running the task (e.g., call \texttt{subprocess.Popen(args).wait()}). When you think you have completed the task, please run the following command: \texttt{<execute\_bash> exit </execute\_bash>}.\\
\hline
\rowcolor{gray!20}
\textbf{Steps:}\\
\hline
\rowcolor{red!10}
\textbf{Create the \texttt{run.sh} file using Python code}\\
\hline
Attempt to create the \texttt{run.sh} file using Python code fails due to a syntax error:
\begin{lstlisting}[numbers=none, backgroundcolor=\color{red!10}]
Running Python code interactively: create_file('/workspace/esm/run.sh')

  File "/opendevin/plugins/jupyter/execute_cli.py", line 34
    if jupyter_pwd := os.environ.get('JUPYTER_PWD'):
                    ^
SyntaxError: invalid syntax
\end{lstlisting}\\
\hline
\rowcolor{orange!10}
\textbf{Task cannot be completed}\\
\hline
The task cannot be completed due to the syntax error in the Python code:
\begin{lstlisting}[numbers=none, backgroundcolor=\color{orange!10}]
I cannot complete the task as requested due to a syntax error in the Python code used to create the run.sh file.
\end{lstlisting}\\
\bottomrule
\end{tabular}
\caption{An example of E4 error. Task execution attempt for Sample sequences design for certain structure. The task could not be completed due to a syntax error in the Python code used to create the \texttt{run.sh} file.}
\label{table:task_error_E4}
\end{table*}

%% file: agent_tables/example_E5.tex
\begin{table*}[!hbtp]
\centering
\begin{tabular}{p{0.98\linewidth}}
\toprule
\hline
\rowcolor{gray!20}
\textbf{Instruction:}\\
\hline
Can you assist me in writing the code to extract 3 designs from the structure \texttt{5YH2.pdb}, specifically from chain B, at a temperature of 1? The output should be saved to the file \texttt{model\_structure.fasta}.\\
\hline
\rowcolor{gray!20}
\textbf{Steps:}\\
\hline
\rowcolor{green!10}
\textbf{Navigate to the repository}\\
\hline

\begin{lstlisting}[numbers=none, language=bash, backgroundcolor=\color{green!10}]
cd /workspace/esm/
\end{lstlisting}\\

\hline
\rowcolor{red!10}
\textbf{Ask the user to create a \texttt{run.sh}}\\
\hline

Now let's create the \texttt{run.sh} script to extract the designs from the structure.
Please create a new file named \texttt{run.sh} in the \texttt{esm} repository and add the following code to it:
\begin{lstlisting}[numbers=none, language=bash, backgroundcolor=\color{red!10}]
#!/bin/bash

python extract_designs.py --structure 5YH2.pdb --chain B --temperature 1 --output model_structure.fasta
\end{lstlisting}\\

\hline
\rowcolor{green!10}
\textbf{Run command}\\
\hline

\begin{lstlisting}[numbers=none, language=bash, backgroundcolor=\color{green!10}]
./run.sh
\end{lstlisting}

\begin{lstlisting}[numbers=none, language=bash, backgroundcolor=\color{orange!10}]
bash: /workspace/esm/run.sh: No such file or directory
ERROR conda.cli.main_run:execute(47): (*@`conda run bash /workspace/esm/run.sh`@*) failed. (See above for error)
\end{lstlisting} \\

\hline
\rowcolor{red!10}
\textbf{Ask the user to create a \texttt{run.sh}}\\
\hline

... \\

\hline
\rowcolor{orange!10}
\textbf{Failed after so many iterations}\\
\hline

\begin{lstlisting}[numbers=none, language=bash, backgroundcolor=\color{orange!10}]
bash: /workspace/esm/run.sh: No such file or directory
ERROR conda.cli.main_run:execute(47): (*@`conda run bash /workspace/esm/run.sh`@*) failed. (See above for error)
\end{lstlisting} \\
\bottomrule
\end{tabular}
\caption{An example of E5 error. Task execution attempt for extracting designs from \texttt{5YH3.pdb}. The LLM agent mistakenly asks the human for help and refuses to do the task by itself.}
\label{table:task_error_E5}
\end{table*}

%% file: github_tables/example_AS.tex
\begin{table*}
\centering
\begin{tabular}{p{0.98\linewidth}}
\toprule

\hline
\rowcolor{gray!20}
\textbf{README:}\\
\hline

As a supplement to the project, an object detection codebase, YOLO. Air has recently been opened. It integrates various attention mechanisms in the object detection algorithm. The code is simple and easy to read. Welcome to play and star!

\textbf{For beginners (like me):}
Recently, I found a problem when reading the paper. Sometimes the core idea of the paper is very simple, and the core code may be just a dozen lines. However, when I open the source code of the author's release, I find that the proposed module is embedded in the task framework such as classification, detection, and segmentation, resulting in redundant code. For me who is not familiar with the specific task framework, it is difficult to find the core code, resulting in some difficulties in understanding the paper and network ideas.

\textbf{For advanced (like you):}
If the basic units conv, FC, and RNN are regarded as small Lego blocks, and the structures transformer and RESNET are regarded as LEGO castles that have been built, the modules provided by this project are LEGO components with complete semantic information. To avoid repeatedly building wheels, scientific researchers should just think about how to use these "LEGO components" to build more colorful works.

\textbf{For proficient (maybe like you):}
Limited capacity, do not like light spraying!!!

\textbf{For All:}
This project aims to realize a code base that can make beginners of deep learning understand and serve scientific research and industrial communities. As fightingcv WeChat official account. The purpose of this project is to achieve Let there be no hard-to-read papers in the world.
(at the same time, we also welcome all scientific researchers to sort out the core code of their work into this project, promote the development of the scientific research community, and indicate the author of the code in README)

...\\

\hline
\rowcolor{gray!20}
\textbf{Oracle Segment:}\\
\hline
..\\
from model.attention.ViP import WeightedPermuteMLP\\
import torch\\
from torch import nn\\
from torch.nn import functional as F\\
...\\

\hline
\rowcolor{gray!20}
\textbf{Instruction:}\\
\hline

I'm planning to utilize the fighting-cv model to complete the attention layers for ViP Attention Usage. Could you provide me with some guidance on accomplishing this task? \\

\hline
\rowcolor{gray!20}
\textbf{Instruction:}\\
\hline

package\_1: ViP \\
sub\_package: WeightedPermuteMLP \\
package\_2: torch \\

\hline
\rowcolor{gray!20}
\textbf{Ground Truth Output:}\\
\hline

from model.attention.ViP import WeightedPermuteMLP\\
import torch\\
from torch import nn\\
from torch.nn import functional as F\\
input=torch.randn(64,8,8,512)\\
seg\_dim=8\\
vip=WeightedPermuteMLP(512,seg\_dim)\\
out=vip(input)\\
print(out.shape)\\

\bottomrule

\end{tabular}
\caption{Example of input-output for \textbf{External-Attention-pytorch} GitHub on \textbf{attention layer} task on \textbf{Attention Usage} domain.
The README URL is \url{https://github.com/xmu-xiaoma666/External-Attention-pytorch/blob/master/README_EN.md}.
The GitHub URL is \url{https://github.com/xmu-xiaoma666/External-Attention-pytorch}.
}
\label{table:example_as}
\end{table*}

%% file: github_tables/example_bert.tex
\begin{table*}
\centering
\begin{tabular}{p{0.98\linewidth}}
\toprule

\hline
\rowcolor{gray!20}
\textbf{README:}\\
\hline
\textbf{BERT}\\

\*\*\*\*\* New March 11th, 2020: Smaller BERT Models \*\*\*\*\* \\

This is a release of 24 smaller BERT models (English only, uncased, trained with WordPiece masking) referenced in Well-Read Students Learn Better: On the Importance of Pre-training Compact Models.\\

...\\

\hline
\rowcolor{gray!20}
\textbf{Oracle Segment:}\\
\hline
..\\
This demo code only pre-trains for a smallnumber of steps (20), but in practice you will probably want to set `num\_train\_steps` to 10000 steps or more. The `max\_seq\_length` and `max\_predictions\_per\_seq` parameters passed to `run\_pretraining.py` must be the same as `create\_pretraining\_data.py`.\\
\\
python run\_pretraining.py\\
\quad\verb|--|input\_file=/tmp/tf\_examples.tfrecord\\
\quad\verb|--|output\_dir=/tmp/pretraining\_output\\
\quad\verb|--|do\_train=True\\
\quad\verb|--|do\_eval=True\\
\quad\verb|--|bert\_config\_file=\$BERT\_BASE\_DIR/bert\_config.json\\
\quad\verb|--|init\_checkpoint=\$BERT\_BASE\_DIR/bert\_model.ckpt\\
...\\

\hline
\rowcolor{gray!20}
\textbf{Instruction:}\\
\hline
Behold, a formidable quest awaits - the pre-training of the unparalleled uncased\_L-24\_H-1024\_A-16 model. Our path to victory lies in configuring the maximum sequence length to a mighty 512, with a pledge to uphold the limit of 30 predictions per sequence. Battling through the treacherous terrain of 10000 steps, we shall march forward, with a stalwart battalion of 32 batch size by our side. But fear not, for we shall brace ourselves with 10000 warmup steps, as we navigate the perilous sea of learning, with a steadfast learning rate of 0.0001. I humbly beseech your assistance, noble comrade, to conjure the code necessary to conquer this heroic endeavor.\\

\hline
\rowcolor{gray!20}
\textbf{Arguments Requirements:}\\
\hline

model: ./model/uncased\_L-12\_H-768\_A-16 \\
train\_batch\_size: 32 \\
max\_seq\_length: 512 \\
num\_train\_steps: 10000 \\
num\_warmup\_steps: 1000 \\
learning\_rate: 0.0001\\

\hline
\rowcolor{gray!20}
\textbf{Ground Truth Output:}\\
\hline

python run\_pretraining.py\\
\quad \verb|--|input\_file=/tmp/tf\_examples.tfrecord\\
\quad \verb|--|output\_dir=/tmp/pretraining\_output\\
\quad \verb|--|do\_train=True\\
\quad \verb|--|do\_eval=True\\   
\quad \verb|--|bert\_config\_file=./model/uncased\_L-24\_H-1024\_A-16/bert\_config.json\\
\quad \verb|--|init\_checkpoint=./model/uncased\_L-24\_H-1024\_A-16/bert\_model.ckpt\\
\quad \verb|--|train\_batch\_size=32\\
\quad \verb|--|max\_seq\_length=512\\
...\\

\bottomrule

\end{tabular}
\caption{Example of input-output for \textbf{bert} GitHub on \textbf{pre-training} task on \textbf{BERT} domain.
The README URL is \url{https://github.com/google-research/bert/blob/master/README.md}.
The GitHub URL is \url{https://github.com/google-research/bert}.
}
\label{table:example_bert}
\end{table*}

%% file: github_tables/example_DGL.tex
\begin{table*}
\centering
\begin{tabular}{p{0.98\linewidth}}
\toprule
\hline
\rowcolor{gray!20}
\textbf{README:}\\
\hline

\textbf{1. DGL Implementation of CorrectAndSmooth}\\
This DGL example implements the GNN model proposed in the paper Combining Label Propagation and Simple Models Out-performs Graph Neural Networks. For the original implementation, see here.\\
Contributor: xnuohz\\
\\
\textbf{2. Requirements}\\
The codebase is implemented in Python 3.7. For version requirement of packages, see below.\\

dgl 0.6.0.post1\\
torch 1.7.0\\
ogb 1.3.0\\
...\\

\hline
\rowcolor{gray!20}
\textbf{Oracle Segment:}\\
\hline
..\\
\textbf{3.1 ogbn-arxiv}\\
Plain MLP + C\&S\\
\\
python main.py\\
\quad--dropout 0.5\\
\\
python main.py\\
\quad\verb|--|pretrain\\
\quad\verb|--|correction-adj DA\\
\quad\verb|--|smoothing-adj AD\\
\quad\verb|--|autoscale\\
...\\


\hline
\rowcolor{gray!20}
\textbf{Instruction:}\\
\hline
...\\
Together, we shall embark on a noble mission to train the illustrious CorrectAndSmooth Model, fortified with a sublime dropout rate of 0.7. Our arduous journey spans 700 epochs, each pulsating with the promise of enlightenment. Alas, I beseech your sage guidance in the ethereal realm of code crafting, to manifest this grand undertaking.\\
\hline
\rowcolor{gray!20}
\textbf{Arguments Requirements:}\\
\hline
dataset: ogbn-arxiv \\
model: mlp \\
dropout: 0.7 \\
epochs: 700 \\
\hline
\rowcolor{gray!20}
\textbf{Ground Truth Output:} \\
\hline
python main.py \\
\quad \verb|--|dataset ogbn-arxiv \\
\quad \verb|--|model mlp \\
\quad \verb|--|dropout 0.7 \\
\quad \verb|--|epochs 700 \\

\bottomrule

\end{tabular}
\caption{Example of input-output for \textbf{DGL} GitHub on \textbf{DGL Implementation of CorrectAndSmooth} task on \textbf{GNN} domain.
The README URL is \url{https://github.com/dmlc/dgl/blob/master/examples/pytorch/correct_and_smooth/README.md}.
The GitHub URL is \url{https://github.com/dmlc/dgl}.
}
\label{table:example_DGL}
\end{table*}

%% file: github_tables/example_ESM.tex
\begin{table*}
\centering
\begin{tabular}{p{0.98\linewidth}}
\toprule
\hline
\rowcolor{gray!20}
\textbf{README:}\\
\hline

\textbf{Evolutionary Scale Modeling}\\

atlas\\
\\
Update April 2023: Code for the two simultaneous preprints on protein design is now released! Code for "Language models generalize beyond natural proteins" is under examples/lm-design/. Code for "A high-level programming language for generative protein design" is under examples/protein-programming-language\\
\\
This repository contains code and pre-trained weights for Transformer protein language models from the Meta Fundamental AI Research Protein Team (FAIR), including our state-of-the-art ESM and ESMFold, as well as MSA Transformer, ESM-1v for predicting variant effects and ESM-IF1 for inverse folding.\\
...\\

\hline
\rowcolor{gray!20}
\textbf{Oracle Segment:}\\
\hline
..\\
The following commands allow the extraction of the final-layer embedding for a FASTA file from the ESM-2 model:\\
esm-extract esm2\_t33\_650M\_UR50D examples/data/some\_proteins.fasta \\examples/data/some\_proteins\_emb\_esm2 \\
\quad \verb|--|repr\_layers 0 32 33 \\
\quad \verb|--|include\\
\\
python scripts/extract.py esm2\_t33\_650M\_UR50D examples/data/some\_proteins.fasta \\examples/data/some\_proteins\_emb\_esm2 \\
\quad \verb|--|repr\_layers 0 32 33 \\
\quad \verb|--|include mean per\_tok \\
\quad \verb|--|A cuda device is optional and will be auto-detected.\\
...\\

\hline
\rowcolor{gray!20}
\textbf{Instruction:}\\
\hline
...\\
Can you assist me in writing the code to extract the 24-layer embedding for a FASTA file named rna.fasta using the esm1v\_t33\_650M\_UR90S\_5 model and save the output?\\
\hline
\rowcolor{gray!20}
\textbf{Arguments Requirements:}\\
\hline
model: esm1v\_t33\_650M\_UR90S\_5 \\
data: rna.fasta \\
layer\_number: 24 \\
layer\_name: repr\_layers \\
\hline
\rowcolor{gray!20}
\textbf{Ground Truth Output:} \\
\hline

python scripts/extract.py esm1v\_t33\_650M\_UR90S\_5 rna.fasta output.embeddings \\
\quad \verb|--|repr\_layers 24 \\
\quad \verb|--|include mean per\_tok \\

\bottomrule

\end{tabular}
\caption{Example of input-output for \textbf{ESM} GitHub on \textbf{Extract ESMFold Structure Prediction Model's Embedding} task on \textbf{molecular} domain.
The README URL is \url{https://github.com/facebookresearch/esm/blob/master/README.md}.
The GitHub URL is \url{https://github.com/facebookresearch/esm}.
}
\label{table:example_esm}
\end{table*}

%% file: github_tables/example_gsa.tex
\begin{table*}
\centering
\begin{tabular}{p{0.98\linewidth}}
\toprule
\hline
\rowcolor{gray!20}
\textbf{README:}\\
\hline

Official PyTorch implementation of Grounding DINO), a stronger open-set object detector. Code is available now!\\
\\
\textbf{Highlight}\\
- Open-Set Detection. Detect everything with language!\\
- High Performancce. COCO zero-shot 52.5 AP (training without COCO data!). COCO fine-tune 63.0 AP.\\
- Flexible. Collaboration with Stable Diffusion for Image Editing.\\
\\
...\\

\hline
\rowcolor{gray!20}
\textbf{Oracle Segment:}\\
\hline
..\\
Demo\\
python demo/inference\_on\_a\_image.py\\
\quad -c /path/to/config\\
\quad -p /path/to/checkpoint\\
\quad -i .asset/cats.png\\
\quad -o outputs/0\\
\quad -t cat ear. [\verb|--|cpu-only] \# open it for cpu mode\\
See the demo/inference\_on\_a\_image.py for more details. \\
...\\

\hline
\rowcolor{gray!20}
\textbf{Instruction:}\\
\hline
...\\
I am interested in utilizing the grounding dino demo for a specific task. The input image path is ground\_segment/GD\_new.json, and I would like the output to be saved in the directory output/cat2002. Additionally, I would like the text condition to be set to right ear of cat. Could you kindly assist me in writing the script to achieve this?\\
\hline
\rowcolor{gray!20}
\textbf{Arguments Requirements:}\\
\hline
i: .asset/cat.jpg \\
o: output/cat2002 \\
t: right ear of cat \\
\hline
\rowcolor{gray!20}
\textbf{Ground Truth Output:} \\
\hline

python demo/inference\_on\_a\_image.py \\
\quad -c model/GroundingDINO\_SwinT\_OGC.py\\
\quad -p model/groundingdino\_swint\_ogc.pth\\
\quad -i .asset/cat.jpg\\
\quad -o output/cat2002\\
\quad -t right ear of cat\\

\bottomrule

\end{tabular}
\caption{Example of input-output for \textbf{Grounded-Segment-Anything} GitHub on \textbf{Grounding DINO demo} task on \textbf{Segment} domain.
The README URL is \url{https://github.com/IDEA-Research/Grounded-Segment-Anything/blob/main/GroundingDINO/README.md}.
The GitHub URL is \url{https://github.com/IDEA-Research/Grounded-Segment-Anything}.
}
\label{table:example_gsa}
\end{table*}

%% file: github_tables/example_IF.tex
\begin{table*}
\centering
\begin{tabular}{p{0.98\linewidth}}
\toprule
\hline
\rowcolor{gray!20}
\textbf{README:}\\
\hline

We introduce DeepFloyd IF, a novel state-of-the-art open-source text-to-image model with a high degree of photorealism and language understanding. DeepFloyd IF is a modular composed of a frozen text encoder and three cascaded pixel diffusion modules: a base model that generates 64x64 px image based on text prompt and two super-resolution models, each designed to generate images of increasing resolution: 256x256 px and 1024x1024 px. All stages of the model utilize a frozen text encoder based on the T5 transformer to extract text embeddings, which are then fed into a UNet architecture enhanced with cross-attention and attention pooling. The result is a highly efficient model that outperforms current state-of-the-art models, achieving a zero-shot FID score of 6.66 on the COCO dataset. Our work underscores the potential of larger UNet architectures in the first stage of cascaded diffusion models and depicts a promising future for text-to-image synthesis.\\
\\


...\\

\hline
\rowcolor{gray!20}
\textbf{Oracle Segment:}\\
\hline
..\\
\textbf{II. Zero-shot Image-to-Image Translation}\\
\\
In Style Transfer mode, the output of your prompt comes out at the style of the support\_pil\_img\\
\\
from deepfloyd\_if.pipelines import style\_transfer\\
\\
result = style\_transfer(\\
t5 = t5, if\_I = if\_I, if\_II = if\_II,\\
...\\


\hline
\rowcolor{gray!20}
\textbf{Instruction:}\\
\hline
...\\
Time to create a visual masterpiece! I am excited to recreate 'image'.jpg into a fascinating rendition of the ink wash style. I am planning to utilize the capabilities of the IF-I-XL-v1.0 model for this endeavor. Can you assist me in crafting the necessary code?\\
\hline
\rowcolor{gray!20}
\textbf{Arguments Requirements:}\\
\hline
model: IF-I-XL-v1.0 \\
argument1: image.jpg \\
argument2: wash \\
\hline
\rowcolor{gray!20}
\textbf{Ground Truth Output:} \\
\hline

from deepfloyd\_if.modules import IFStageI, IFStageII, StableStageIII\\
from deepfloyd\_if.modules.t5 import T5Embedder\\
device = 'cuda:1'\\
if\_I = IFStageI('IF-I-XL-v1.0', device=device)\\
if\_II = IFStageII('IF-II-L-v1.0', device=device)\\
if\_III = StableStageIII('stable-diffusion-x4-upscaler', device=device)\\
t5 = T5Embedder(device='cpu')\\
\\
style\_prompt = a captivating ink wash style\\
image\_path = image.jpg\\
...\\
\bottomrule

\end{tabular}
\caption{Example of input-output for \textbf{IF} GitHub on \textbf{Zero-shot Image-to-Image Translation} task on \textbf{Image and Text} domain.
The README URL is \url{https://github.com/deep-floyd/IF/blob/develop/README.md}.
The GitHub URL is \url{https://github.com/deep-floyd/if}.
}
\label{table:example_if}
\end{table*}

%% file: github_tables/example_lavis.tex
\begin{table*}
\centering
\begin{tabular}{p{0.98\linewidth}}
\toprule

\hline
\rowcolor{gray!20}
\textbf{README:}\\
\hline
\textbf{Lavis}\\

Lavis - A Library for Language-Vision Intelligence\\
What's New: \\
A simple, yet effective, cross-modality framework built atop frozen LLMs that allows the integration of various modalities (image, video, audio, 3D) without extensive modality-specific customization.\\
Technical Report and Citing LAVIS: \\

...\\

\hline
\rowcolor{gray!20}
\textbf{Oracle Segment:}\\
\hline
...\\
how to use models in LAVIS to perform inference on example data. We first load a sample image from local.\\
import torch\\
from PIL import Image\\
\# setup device to use\\
device = torch.device("cuda" if torch.cuda.is\_available() else "cpu")\\
\# load sample image\\
raw\_image = Image.open(merlion.png).convert(RGB)\\
This example image shows Merlion park (source), a landmark in Singapore.\\
\# Image Captioning\\
In this example, we use the BLIP model to generate a caption for the image.\\
...\\

\hline
\rowcolor{gray!20}
\textbf{Instruction:}\\
\hline
There is a task in front of me currently, which is the Image Captioning task. I ask for your kind help in moving forward with this task.\\

\hline
\rowcolor{gray!20}
\textbf{Arguments Requirements:}\\
\hline

image: image.jpg \\

\hline
\rowcolor{gray!20}
\textbf{Ground Truth Output:}\\
\hline
import torch \\

from PIL import Image \\
from lavis.models import load\_model\_and\_preprocess \\

device = torch.device('cuda:1' if torch.cuda.is\_available() else 'cpu')\\
model, vis\_processors, \_ = load\_model\_and\_preprocess(name='blip\_caption', \\
model\_type='base\_coco', is\_eval=True, device=device)\\
raw\_image = Image.open('image.jpg').convert('RGB')\\
preprocessed\_image = vis\_processors['eval'](raw\_image).unsqueeze(0).to(device)\\
output = model.generate(\{'image': preprocessed\_image\})\\
caption = output['captions'][0]['text']\\
print(caption)\\
\bottomrule

\end{tabular}
\caption{Example of input-output for \textbf{Lavis} GitHub on \textbf{Image Captioning} task on \textbf{Multimodal Image and Text} domain.
The README URL is \url{https://github.com/salesforce/LAVIS/blob/main/README.md}.
The GitHub URL is \url{https://github.com/salesforce/lavis}.
}
\label{table:example_lavis}
\end{table*}

%% file: github_tables/example_learning3D.tex
\begin{table*}
\centering
\begin{tabular}{p{0.98\linewidth}}
\toprule

\hline
\rowcolor{gray!20}
\textbf{README:}\\
\hline

\textbf{Learning3D: A Modern Library for Deep Learning on 3D Point Clouds Data.}\\
Learning3D is an open-source library that supports the development of deep learning algorithms that deal with 3D data. The Learning3D exposes a set of state of art deep neural networks in python. A modular code has been provided for further development. We welcome contributions from the open-source community. \\
\textbf{Available Computer Vision Algorithms in Learning3D}\\
...\\

\hline
\rowcolor{gray!20}
\textbf{Oracle Segment:}\\
\hline
...\\
examples/test\_dcp.py Learning3D is an open-source library that supports the development of deep learning algorithms that deal with 3D data. The Learning3D exposes a set of state of art deep neural networks in python\\
python test\_dcp.py\\
\quad\verb|--|num\_points 128 \\
\quad\verb|--|j 12 \\
\quad\verb|--|symfn max\\
...\\

\hline
\rowcolor{gray!20}
\textbf{Instruction:}\\
\hline
I am interested in conducting a test using the dcp model. Specifically, I would like to set the parameters as follows: the test mode should be selected, the model should be set to dcp, the number of points should be 512, the number of data loading workers should be -j 8, and the symmetric function should be set to --symfn max. Could you please assist me in writing the code or script necessary to carry out this test?\\

\hline
\rowcolor{gray!20}
\textbf{Arguments Requirements:}\\
\hline
number of points: 512\\
number of data loading workers: 8\\
symmetric function: max\\

\hline
\rowcolor{gray!20}
\textbf{Ground Truth Output:}\\
\hline
python test\_dcp.py\\
\quad\verb|--|num\_points 512 \\
\quad\verb|--|j 8 \\
\quad\verb|--|symfn max\\
\bottomrule

\end{tabular}
\caption{Example of input-output for \textbf{Learning3D} GitHub on \textbf{Test dcp model} task on \textbf{3D} domain.
The README URL is \url{https://github.com/vinits5/learning3d/blob/master/README.md}.
The GitHub URL is \url{https://github.com/vinits5/learning3d}.
}
\label{table:example_learning3D}
\end{table*}

%% file: github_tables/example_MedicalZoo.tex
\begin{table*}
\centering
\begin{tabular}{p{0.98\linewidth}}
\toprule

\hline
\rowcolor{gray!20}
\textbf{README:}\\
\hline

\textbf{MusicBERT}\\
Basics\\
All models accept two parameters: a) the input the channels (in\_channels), and b) the segmentation classes (classes) and produce un-normalized outputs\\
All losses accept as input the prediction in 5D shape of [batch,classes,dim\_1,dim\_2,dim\_3] and the target in 4D target shape of [batch, dim\_1, dim\_2, dim\_3]. It is converted to one-hot inside the loss function for consistency reasons. \\
Furthermore the normalization of the predictions is handled here. Dice-based losses return the scalar loss for backward(), and the prediction per channels in numpy to track training progress.\\

...\\

\hline
\rowcolor{gray!20}
\textbf{Oracle Segment:}\\
\hline
Usage\\
How to train your model\\
For Iseg-2017 :\\
python ./examples/train\_iseg2017\_new.py\\
\quad\verb|--|args\\
For MR brains 2018 (4 classes)\\
python ./examples/train\_mrbrains\_4\_classes.py\\
\quad\verb|--|args\\
For MR brains 2018 (8 classes)\\
python ./examples/train\_mrbrains\_9\_classes.py \\
\quad\verb|--|args\\
For MICCAI 2019 Gleason Challenge\\
python ./examples/test\_miccai\_2019.py \\
\quad\verb|--|args\\
The arguments that you can modify are extensively listed in the manual.\\

\hline
\rowcolor{gray!20}
\textbf{Instruction:}\\
\hline
I'm seeking assistance in writing a piece of code that can successfully train a model for the 'Iseg 2017 Task'. The model in question is 'RESNET3DVAE' and I require the learning rate to be set to '1e-3'. It is also crucial that the training samples are set to '10'. Lastly, use 'sgd' as the optimizer. Could you kindly help out in creating this algorithm?\\

\hline
\rowcolor{gray!20}
\textbf{Arguments Requirements:}\\
\hline
lr: 1e-3\\
samples\_train: 10\\
model: RESNET3DVAE\\
soptimizer: sg\\

\hline
\rowcolor{gray!20}
\textbf{Ground Truth Output:}\\
\hline
python ./examples/train\_iseg2017\_new.py\\
\quad\verb|--|lr 1e-3 \\
\quad\verb|--|samples\_train 10 \\
\quad\verb|--|model RESNET3DVAE \\
\quad\verb|--|opt sgd\\
\bottomrule

\end{tabular}
\caption{Example of input-output for \textbf{MedicalZoo} GitHub on \textbf{Iseg-2017} task on \textbf{Medical} domain.
The README URL is \url{https://github.com/black0017/MedicalZooPytorch/blob/master/manual/README.md}.
The GitHub URL is \url{https://github.com/black0017/MedicalZooPytorch}.
}
\label{table:example_medicalzoo}
\end{table*}

%% file: github_tables/example_muzic.tex
\begin{table*}
\centering
\begin{tabular}{p{0.98\linewidth}}
\toprule

\hline
\rowcolor{gray!20}
\textbf{README:}\\
\hline

\textbf{MusicBERT}\\
MusicBERT: Symbolic Music Understanding with Large-Scale Pre-Training, by Mingliang Zeng, Xu Tan, Rui Wang, Zeqian Ju, Tao Qin, Tie-Yan Liu, ACL 2021, is a large-scale pre-trained model for symbolic music understanding. It has several mechanisms including OctupleMIDI encoding and bar-level masking strategy that are specifically designed for symbolic music data, and achieves state-of-the-art accuracy on several music understanding tasks, including melody completion, accompaniment suggestion, genre classification, and style classification.\\
Projects using MusicBERT:\\

midiformers: a customized MIDI music remixing tool with easy interface for users.\\

\textbf{1. Preparing datasets}\\

\textbf{1.1 Pre-training datasets}\\

Prepare\\
tar -xzvf lmd\_full.tar.gz\\
zip -r lmd\_full.zip lmd\_full\\

Run the dataset processing script. (`preprocess.py`)\\

python -u preprocess.py\\

The script should prompt you to input the path of the midi zip and the path for OctupleMIDI output.\\

...\\

\hline
\rowcolor{gray!20}
\textbf{Oracle Segment:}\\
\hline
Pre-training bash train\_mask.sh lmd\_full small Download our pre-trained checkpoints here: small and base, and save in the checkpoints folder. (a newer version of fairseq is needed for using provided checkpoints: see issue-37 or issue-45)\\

\hline
\rowcolor{gray!20}
\textbf{Instruction:}\\
\hline
I am interested in conducting a test using the dcp model. Specifically, I would like to set the parameters as follows: the test mode should be selected, the model should be set to dcp, the number of points should be 512, the number of data loading workers should be -j 8, and the symmetric function should be set to --symfn max. Could you please assist me in writing the code or script necessary to carry out this test?\\

\hline
\rowcolor{gray!20}
\textbf{Arguments Requirements:}\\
\hline
bash: train\_mask.sh\\
dataset: lmd\_full\\
checkpoint: small\\

\hline
\rowcolor{gray!20}
\textbf{Ground Truth Output:}\\
\hline
bash train\_mask.sh lmd\_full small\\
\bottomrule

\end{tabular}
\caption{Example of input-output for \textbf{Muzic} GitHub on \textbf{Pre-training model} task on \textbf{Music} domain.
The README URL is \url{https://github.com/microsoft/muzic/blob/main/musicbert/README.md}.
The GitHub URL is \url{https://github.com/microsoft/muzic}.
}
\label{table:example_muzic}
\end{table*}

%% file: github_tables/example_openCLIP.tex
\begin{table*}
\centering
\begin{tabular}{p{0.98\linewidth}}
\toprule

\hline
\rowcolor{gray!20}
\textbf{README:}\\
\hline

\textbf{OpenCLIP}\\
Welcome to an open-source implementation of OpenAI's CLIP(Contrastive Language-Image Pre-training).

Using this codebase, we have trained several models on a variety of data sources and compute budgets, ranging from small-scale experiments to larger runs including models trained on datasets such as LAION-400M, LAION-2B and DataComp-1B.\\
Many of our models and their scaling properties are studied in detail in the paper reproducible scaling laws for contrastive language-image learning.\\
Some of our best models and their zero-shot ImageNet-1k accuracy are shown below, along with the ViT-L model trained by OpenAI. 
We provide more details about our full collection of pretrained models here, and zero-shot results for 38 datasets here.\\

midiformers: a customized MIDI music remixing tool with easy interface for users.\\
...\\

\hline
\rowcolor{gray!20}
\textbf{Oracle Segment:}\\
\hline
Pre-training bash train\_mask.sh lmd\_full small Download our pre-trained checkpoints here: small and base, and save in the checkpoints folder. (a newer version of fairseq is needed for using provided checkpoints: see issue-37 or issue-45)\\

\hline
\rowcolor{gray!20}
\textbf{Instruction:}\\
\hline
I'm hoping you could assist me with a task. I need to sort a collection of images located at the directory '. /image'. For this purpose, I'll be using the 'ViT-B-32' model along with the pretraining model 'commonpool\_s\_laion\_s13m\_b4k'. Would you be able to provide instructions or steps on how to accomplish this task?\\

\hline
\rowcolor{gray!20}
\textbf{Arguments Requirements:}\\
\hline
model: ViT-B-32\\
pretraining\_model: commonpool\_s\_laion\_s13m\_b4k\\
path: ./image\\

\hline
\rowcolor{gray!20}
\textbf{Ground Truth Output:}\\
\hline
import os\\
import numpy as np\\
import torch\\
import open\_clip\\
open\_clip.list\_pretrained()\\
model, \_, preprocess = \\
open\_clip.create\_model\_and\_transforms('ViT-B-32', pretrained='commonpool\_s\_laion\_s13m\_b4k')\\
model.eval()\\
context\_length = model.context\_length\\
vocab\_size = model.vocab\_size\\
... \\

\bottomrule

\end{tabular}
\caption{Example of input-output for \textbf{OpenCLIP} GitHub on \textbf{Image classification} task on \textbf{Image and Text} domain.
The README URL is \url{https://github.com/mlfoundations/open_clip/blob/main/README.md}.
The GitHub URL is \url{https://github.com/mlfoundations/open_clip}.
}
\label{table:example_openCLIP}
\end{table*}

%% file: github_tables/example_PyGAN.tex
\begin{table*}
\centering
\begin{tabular}{p{0.98\linewidth}}
\toprule

\hline
\rowcolor{gray!20}
\textbf{README:}\\
\hline

\textbf{MusicBERT}\\

Collection of PyTorch implementations of Generative Adversarial Network varieties presented in research papers. Model architectures will not always mirror the ones proposed in the papers, but I have chosen to focus on getting the core ideas covered instead of getting every layer configuration right. Contributions and suggestions of GANs to implement are very welcomed.\\
Table of Contents:\\
\quad Installation\\
\quad Implementations\\
\quad Auxiliary Classifier GAN\\
\quad Adversarial Autoencoder\\
...\\

\hline
\rowcolor{gray!20}
\textbf{Oracle Segment:}\\
\hline
...\\
Energy-Based GAN\\
Among them, we show one instantiation of EBGAN framework as using an auto-encoder architecture, with the energy being the reconstruction error, in place of the discriminator. We show that this form of EBGAN exhibits more stable behavior than regular GANs during training. We also show that a single-scale architecture can be trained to generate high-resolution images.  \\
Run Example \\
\$ cd implementations/ebgan/  \\
\$ python3 ebgan.py \\

\hline
\rowcolor{gray!20}
\textbf{Instruction:}\\
\hline
I have a task to work with the Energy-Based GAN model. The learning rate for this task needs to be set at 0.0001, the number of training epochs should be defined as 100, and the batch size should be fixed at 16. Furthermore, I want the image size to be set at 128. Can you please assist me in framing the script to facilitate this?\\

\hline
\rowcolor{gray!20}
\textbf{Arguments Requirements:}\\
\hline
lr: 0.0001\\
n\_epochs: 100\\
batch\_size: 16\\
img\_size: 128\\
model: ebgan\\

\hline
\rowcolor{gray!20}
\textbf{Ground Truth Output:}\\
\hline
python3 ebgan.py \\
\quad\verb|--|lr 0.0001\\
\quad\verb|--|n\_epochs 100\\
\quad\verb|--|batch\_size 16 \\
\quad\verb|--|mg\_size 128\\
\bottomrule

\end{tabular}
\caption{Example of input-output for \textbf{pyGAN} GitHub on \textbf{Energy-Based GAN} task on \textbf{images-many-GANs} domain.
The README URL is \url{https://github.com/eriklindernoren/PyTorch-GAN/blob/master/README.md}.
The GitHub URL is \url{https://github.com/eriklindernoren/PyTorch-GAN}.
}
\label{table:example_pyGAN}
\end{table*}

%% file: github_tables/example_PyIM.tex
\begin{table*}
\centering
\begin{tabular}{p{0.98\linewidth}}
\toprule
\hline
\rowcolor{gray!20}
\textbf{README:}\\
\hline

\textbf{PyTorch Image Models}\\
...\\
\textbf{What's new}\\
...\\
\textbf{Introduction}\\
PyTorch Image Models (timm) is a collection of image models, layers, utilities, optimizers, schedulers, data-loaders / augmentations, and reference training / validation scripts that aim to pull together a wide variety of SOTA models with ability to reproduce ImageNet training results.\\
...\\

\hline
\rowcolor{gray!20}
\textbf{Oracle Segment:}\\
\hline
..\\

Existing method of changing patch\_size (resize pretrained patch\_embed weights once) on creation still works.
\\
\textbf{Example validation cmd} \\
python validate.py /imagenet \\
\quad\verb|--|model vit\_base\_patch16\_224\\
\quad\verb|--|amp\\
\quad\verb|--|amp-dtype bfloat16\\
\quad\verb|--|img-size 255\\
\quad\verb|--|crop-pct 1.0\\
\quad\verb|--|model-kwargs dynamic\_img\_size=True dyamic\_img\_pad=True\\

...\\


\hline
\rowcolor{gray!20}
\textbf{Instruction:}\\
\hline
I am interested in performing the task of resizing the image or window. For this purpose, I would like to utilize the model vit\_base\_patch16\_224. Additionally, it would be helpful to set the amp-dtype to bfloat16. Moreover, I would like to specify the image size as 255 and the crop percentage as 1.0. To ensure flexibility, I would like to enable dynamic image size and dynamic image padding. Could you kindly assist me in creating the code or script to accomplish this objective?\\
\hline
\rowcolor{gray!20}
\textbf{Arguments Requirements:}\\
\hline
model: vit\_base\_patch16\_224\\
amp-dtype: bfloat16\\
img-size: 255\\
crop-pct: 1.0\\
dynamic\_img\_size: True\\
dyamic\_img\_pad: True\\
\hline
\rowcolor{gray!20}
\textbf{Ground Truth Output:} \\
\hline

python validate.py /imagenet \\
\quad \verb|--|model vit\_base\_patch16\_224\\
\quad \verb|--|amp\\
\quad \verb|--|amp-dtype bfloat16\\
\quad \verb|--|img-size 255\\
\quad \verb|--|crop-pct 1.0\\
\quad \verb|--|model-kwargs dynamic\_img\_size=True\\

\bottomrule

\end{tabular}

\caption{Example of input-output for \textbf{PyIM} GitHub on \textbf{PyIM Implementation of Resize The Image/Window} task on \textbf{Image} domain. The README URL is \url{https://github.com/huggingface/pytorch-image-models/blob/main/README.md}. The GitHub URL is \url{https://github.com/huggingface/pytorch-image-models}.}
\label{table:example_PyIM}
\end{table*}

%% file: github_tables/example_SD.tex
\begin{table*}
\centering
\begin{tabular}{p{0.98\linewidth}}
\toprule
\hline
\rowcolor{gray!20}
\textbf{README:}\\
\hline

\textbf{Stable Diffusion Version 2}\\
This repository contains Stable Diffusion models trained from scratch and will be continuously updated with new checkpoints. The following list provides an overview of all currently available models. More coming soon.\\
...\\
\textbf{Requirements}\\
You can update an existing latent diffusion environment by running.\\
...\\

\hline
\rowcolor{gray!20}
\textbf{Oracle Segment:}\\
\hline
...\\

We provide the configs for the SD2-v (768px) and SD2-base (512px) model. First, download the weights for SD2.1-v and SD2.1-base. To sample from the SD2.1-v model, run the following:\\
python scripts/txt2img.py \\
\quad \verb|--|prompt "a professional photograph of an astronaut riding a horse" \\
\quad \verb|--|ckpt <path/to/768model.ckpt\verb|\|> \\
\quad \verb|--|config configs/stable-diffusion/v2-inference-v.yaml \\
\quad \verb|--|H 768 \\
\quad \verb|--|W 768  \\
or try out the Web Demo: Hugging Face Spaces.\\

...\\


\hline
\rowcolor{gray!20}
\textbf{Instruction:}\\
\hline
...\\ 
For the task of generating an image from text, I need your assistance in writing the code. We'll be using the scripts/txt2img.py script along with the SD2.1-v model. Ensure that the model checkpoint file is located at ...... As we want to generate a high-quality image, set the number of sampling steps to 20. The prompt to generate the image is "a professional photograph of an astronaut riding a horse" and we only need one iteration of the generation process. Can you help me write the code to accomplish this task?\\
\hline
\rowcolor{gray!20}
\textbf{Arguments Requirements:}\\
\hline
repeat: 1\\
config: "configs/stable-diffusion/v2-inference-v.yaml"\\
ckpt: "ckpt/SD2\_1\_v\_model.ckpt"\\
prompt: "a professional photograph of an astronaut riding a horse"\\
precision: full\\
steps: 20\\
seed: 2048\\
\hline
\rowcolor{gray!20}
\textbf{Ground Truth Output:} \\
\hline

python scripts/txt2img.py \\
\quad \verb|--|prompt "a professional photograph of an astronaut riding a horse"\\
\quad \verb|--|ckpt ckpt/SD2\_1\_v\_model.ckpt \\
\quad \verb|--|config configs/stable-diffusion/v2-inference-v.yaml \\
\quad \verb|--|H 768 \\
\quad \verb|--|W 768 \\
\quad \verb|--|seed 2048 \\
\quad \verb|--|precision full \\
\quad \verb|--|steps 20 \\
\quad \verb|--|repeat 1 \\

\bottomrule

\end{tabular}

\caption{Example of input-output for \textbf{SD} GitHub on \textbf{SD Implementation of Text-to-Image} task on \textbf{Stable Diffusion} domain. The README URL is \url{https://github.com/Stability-AI/stablediffusion/blob/main/README.md}. The GitHub URL is \url{https://github.com/Stability-AI/stablediffusion}.}
\label{table:example_SD}
\end{table*}

%% file: github_tables/example_TC.tex
\begin{table*}
\centering
\begin{tabular}{p{0.98\linewidth}}
\toprule
\hline
\rowcolor{gray!20}
\textbf{README:}\\
\hline

\textbf{Text Classification}\\
The purpose of this repository is to explore text classification methods in NLP with deep learning.\\
...\\
\textbf{Usage:}\\
 1.model is in xxx\_model.py\\
 2.run python xxx\_train.py to train the model\\
...\\

\hline
\rowcolor{gray!20}
\textbf{Oracle Segment:}\\
\hline


it learn represenation of each word in the sentence or document with left side context and right side context:\\
representation current word=[left\_side\_context\_vector,current\_word\_embedding,right\_side\_context\_vecotor].\\
for left side context, it use a recurrent structure, a no-linearity transfrom of previous word and left side previous context; similarly to right side context.check: p71\_TextRCNN\_model.py\\



\hline
\rowcolor{gray!20}
\textbf{Instruction:}\\
\hline
I am looking to utilize the TextRCNN model for a particular task. In the course of executing this task, I would like to fix the learning rate at 0.00001, the number of training epochs at 300, and set my batch size to 16. Are you in a position to assist me in creating the appropriate coding syntax for this purpose?\\
\hline
\rowcolor{gray!20}
\textbf{Arguments Requirements:}\\
\hline
model: TextRCNN\\
learning\_rate: 0.00001\\
num\_epochs: 300\\
batch\_size: 16\\
\hline
\rowcolor{gray!20}
\textbf{Ground Truth Output:} \\
\hline

python3 a04\_TextRCNN/p71\_TextRCNN\_train.py \\
\quad \verb|--|num\_epochs 300\\
\quad \verb|--|batch\_size 16\\
\quad \verb|--|lr 0.00001\\

\bottomrule

\end{tabular}

\caption{Example of input-output for \textbf{TC} GitHub on \textbf{TC Implementation of TextRCNN} task on \textbf{texts} domain. The README URL is \url{https://github.com/brightmart/text_classification/blob/master/README.md}. The GitHub URL is \url{https://github.com/brightmart/text_classification}.}
\label{table:example_TC}
\end{table*}

%% file: github_tables/example_tensor2.tex
\begin{table*}
\centering
\begin{tabular}{p{0.98\linewidth}}
\toprule
\hline
\rowcolor{gray!20}
\textbf{README:}\\
\hline

\textbf{Tensor2Tensor}\\
Tensor2Tensor, or T2T for short, is a library of deep learning models and datasets designed to make deep learning more accessible and accelerate ML research.\\
...\\
\textbf{Quick Start}\\
This iPython notebook explains T2T and runs in your browser using a free VM from Google, no installation needed. Alternatively, here is a one-command version that installs T2T, downloads MNIST, trains a model and evaluates it:\\
...\\

\hline
\rowcolor{gray!20}
\textbf{Oracle Segment:}\\
\hline

...\\
Test the quick-start on a Workspace's Terminal with this command\\
t2t-trainer --generate\_data\\
\quad\verb|--|data\_dir=./t2t\_data\\
\quad\verb|--|output\_dir=./t2t\_train/mnist\\
\quad\verb|--|problem=image\_mnist\\
\quad\verb|--|model=shake\_shake\\
\quad\verb|--|hparams\_set=shake\_shake\_quick\\
\quad\verb|--|train\_steps=1000\\
\quad\verb|--|eval\_steps=100\\
Note: Ensure compliance with the FloydHub Terms of Service.\\

...\\


\hline
\rowcolor{gray!20}
\textbf{Instruction:}\\
\hline
I'd like assistance in trying to make sense of mathematical language. Any help you can give would be highly appreciated, in comprehending the intricate nature of mathematical terminologies and symbols.\\
\hline
\rowcolor{gray!20}
\textbf{Ground Truth Output:} \\
\hline

t2t-trainer --generate\_data\\
\quad\verb|--|data\_dir=~/t2t\_data \\
\quad\verb|--|output\_dir=~/t2t\_train/mathematical\_language\\
\quad\verb|--|problem=algorithmic\_math\_two\_variables \\
\quad\verb|--|model=transformer \\
\quad\verb|--|hparams\_set=transformer\_tiny \\
\quad\verb|--|train\_steps=1000 \\
\quad\verb|--|eval\_steps=100 \\

\bottomrule

\end{tabular}

\caption{Example of input-output for \textbf{tensor2} GitHub on \textbf{tensor2 Implementation of Mathematical Language Understanding} task on \textbf{Text} domain. The README URL is \url{https://github.com/tensorflow/tensor2tensor/blob/master/README.md}. The GitHub URL is \url{https://github.com/tensorflow/tensor2tensor}.}
\label{table:example_tensor2}
\end{table*}

%% file: github_tables/example_tsl.tex
\begin{table*}
\centering
\begin{tabular}{p{0.98\linewidth}}
\toprule
\hline
\rowcolor{gray!20}
\textbf{README:}\\
\hline

\textbf{Time Series Library (TSlib)}\\
TSlib is an open-source library for deep learning researchers, especially for deep time series analysis.\\
We provide a neat code base to evaluate advanced deep time series models or develop your model, which covers five mainstream tasks: long- and short-term forecasting, imputation, anomaly detection, and classification.\\
...\\
\textbf{Usage}\\
1.Install Python 3.8. For convenience, execute the following command.\\
pip install -r requirements.txt\\
...\\

\hline
\rowcolor{gray!20}
\textbf{Oracle Segment:}\\
\hline

...\\
Train and evaluate model. We provide the experiment scripts for all benchmarks under the folder ./scripts/. You can reproduce the experiment results as the following examples:\\
\textbf{long-term forecast}\\
bash\ ./scripts/long\_term\_forecast/ETT\_script/TimesNet\_ETTh1.sh\\
\textbf{short-term forecast}\\
bash\ ./scripts/short\_term\_forecast/TimesNet\_M4.sh\\
...\\



\hline
\rowcolor{gray!20}
\textbf{Instruction:}\\
\hline
I'm aiming to conduct a performance evaluation of my FEDformer model which is designated for a classification task. I would like to utilize subset: FaceDetection of the UEA benchmark dataset for this evaluation. The patience parameter needs to be set to 10, with the train epochs set to 16, and batch size adjusted to 100. Could you kindly provide me the code necessary to accomplish this?
\\
\hline
\rowcolor{gray!20}
\textbf{Arguments Requirements:}\\
\hline
task\_name: classification\\
model: FEDformer\\
data: UEA\\
model\_id: FaceDetection\\
patience: 10\\
batch\_size: 16\\
train\_epochs: 100\\
\hline
\rowcolor{gray!20}
\textbf{Ground Truth Output:} \\
\hline

python -u run.py \\
\quad \verb|--|task\_name classification \\
\quad \verb|--|model FEDformer \\
\quad \verb|--|data UEA \\
\quad \verb|--|e\_layers 3 \\
\quad \verb|--|batch\_size 16 \\
\quad \verb|--|d\_model 128 \\
\quad \verb|--|d\_ff 256 \\
\quad \verb|--|top\_k 3 \\
\quad \verb|--|train\_epochs 100 \\
\quad \verb|--|patience 10 \\
...\\

\bottomrule

\end{tabular}

\caption{Example of input-output for \textbf{tsl} GitHub on \textbf{tsl Implementation of classification} task on \textbf{Time-series} domain. The README URL is \url{https://github.com/thuml/Time-Series-Library/blob/main/README.md}. The GitHub URL is \url{https://github.com/thuml/Time-Series-Library}.}
\label{table:example_tsl}
\end{table*}

%% file: github_tables/example_vid2vid.tex
\begin{table*}
\centering
\begin{tabular}{p{0.98\linewidth}}
\toprule
\hline
\rowcolor{gray!20}
\textbf{README:}\\
\hline

\textbf{vid2vid}\\
...\\
Pytorch implementation for high-resolution (e.g., 2048x1024) photorealistic video-to-video translation. It can be used for turning semantic label maps into photo-realistic videos, synthesizing people talking from edge maps, or generating human motions from poses. The core of video-to-video translation is image-to-image translation. Some of our work in that space can be found in pix2pixHD and SPADE.\\
...\\
\textbf{Getting Started}\\
...\\

\hline
\rowcolor{gray!20}
\textbf{Oracle Segment:}\\
\hline

...\\
To test the model (bash ./scripts/face/test\_512.sh):
\#!./scripts/face/test\_512.sh\\
python test.py \\
\quad \verb|--|name edge2face\_512 \\
\quad \verb|--|dataroot datasets/face/ \\
\quad \verb|--|dataset\_mode face \\
\quad \verb|--|input\_nc 15 \\
\quad \verb|--|loadSize 512 \\
\quad \verb|--|use\_single\_G\\
The test results will be saved in: ./results/edge2face\_512/test\_latest/.\\



\hline
\rowcolor{gray!20}
\textbf{Instruction:}\\
\hline
We have the task of training the edge2face model on a face dataset. Let's set the resolution size this time to 256. Can you provide assistance in drafting the applicable coding structure?
\\
\hline
\rowcolor{gray!20}
\textbf{Arguments Requirements:}\\
\hline
model: edge2face\\
dataset: face\\
resolusion size: 256\\
argument: g1\\
\hline
\rowcolor{gray!20}
\textbf{Ground Truth Output:} \\
\hline

python train.py \\
\quad \verb|--|name edge2face\_256\_g1 \\
\quad \verb|--|dataroot datasets/face/\\
\quad \verb|--|dataset\_mode face \\
\quad \verb|--|input\_nc 15 \\
\quad \verb|--|loadSize 256 \\
\quad \verb|--|ngf 64 \\
\quad \verb|--|max\_frames\_per\_gpu 6 \\
\quad \verb|--|n\_frames\_total 12 \\
\quad \verb|--|niter 20 \\
\quad \verb|--|niter\_decay 20\\

\bottomrule

\end{tabular}

\caption{Example of input-output for \textbf{vid2vid} GitHub on \textbf{vid2vid Implementation of Edge to Face Training} task on \textbf{Videos} domain. The README URL is \url{https://github.com/NVIDIA/vid2vid/blob/master/README.md}. The GitHub URL is \url{https://github.com/NVIDIA/vid2vid}.}
\label{table:example_vid2vid}
\end{table*}